\documentclass{article}

\usepackage[utf8]{inputenc} % allow utf-8 input
\usepackage[T1]{fontenc}    % use 8-bit T1 fonts
\usepackage{url}            % simple URL typesetting
\usepackage{booktabs}       % professional-quality tables
\usepackage{amsfonts}       % blackboard math symbols
\usepackage{nicefrac}       % compact symbols for 1/2, etc.
\usepackage{microtype}      % microtypography
\usepackage{xcolor}         % colors
\usepackage{xspace}
\usepackage{graphicx}
\usepackage{subcaption}
\usepackage[most]{tcolorbox}
\usepackage{enumitem}
\usepackage{pifont}
\usepackage{multirow}
\usepackage{bbm}
\usepackage[dvipsnames]{xcolor}
\usepackage[normalem]{ulem}

\usepackage{pgfplots}
\usepgfplotslibrary{groupplots,fillbetween}
\usetikzlibrary{calc}
\usetikzlibrary{arrows.meta,patterns,backgrounds,decorations.pathreplacing,calc,positioning,shapes.geometric}
\pgfplotsset{compat=1.18}

\newcommand{\ie}{\textit{i.e.\xspace}}
\newcommand{\eg}{\textit{e.g.\xspace}}

\newtcolorbox{widebox}[2][]{
    enhanced jigsaw,
    breakable,
    width=\textwidth,
    colback=white,
    title={#2},
    float*=ht,      
    #1
}

\definecolor{colorPrompt}{HTML}{009E73}

\DeclareCaptionType[within=none]{promptbox}[Prompt][List of prompts]

\newtcolorbox{prompt}[4][]{
  breakable,
  colback=colorPrompt!5!white,
  colframe=colorPrompt!75!black,
  fonttitle=\bfseries\small,
  fontupper=\small,
  title={#2},
  after={\captionsetup{type=promptbox}\captionof{promptbox}{#3}\label{#4}\vspace{3mm}},
  #1
}

\newcommand{\ours}{\texttt{PRE-ACT}\xspace}

\definecolor{myblue}{rgb}{0.19, 0.55, 0.91}
\definecolor{myred}{rgb}{0.82, 0.1, 0.26}
\definecolor{MyGreen}{RGB}{0, 104, 55} %{0, 180, 0}
\definecolor{MyRed}{RGB}{248, 3, 7} %{180, 0, 0}
\definecolor{MyYellow}{RGB}{180, 180, 0} 
\newcommand{\cmark}{{\textcolor{MyGreen}{\ding{51}}}}%
\newcommand{\xmark}{{\textcolor{MyRed}{\ding{55}}}}%

\definecolor{mplmagenta}{RGB}{255,0,255}
\definecolor{mplcyan}{RGB}{0,255,255}
\definecolor{mpllime}{RGB}{0,255,0}

\definecolor{dodgerblue}{RGB}{30,144,255}

\newcommand{\bs}{\mathbf{s}}

\newcommand{\bz}{\mathbf{z}}

\newcommand{\cZ}{\mathcal{Z}}

\newcommand{\figref}[1]{\Fig~\ref{#1}}
\newcommand{\appref}[1]{\App~\ref{#1}}
\newcommand{\secref}[1]{Section~\ref{#1}}

\newcommand{\eqnref}[1]{Eq.~\eqref{#1}}
\newcommand{\tabref}[1]{Table~\ref{#1}}

\makeatletter
\DeclareRobustCommand\onedot{\futurelet\@let@token\@onedot}
\def\@onedot{\ifx\@let@token.\else.\null\fi\xspace}
\def\eg{e.g\onedot} 
\def\ie{i.e\onedot}

\def\Fig{Fig\onedot}   
\def\App{App\onedot}
\makeatother

\newcommand{\xdownarrow}[1]{%
  {\left\downarrow\vbox to #1{}\right.\kern-\nulldelimiterspace}
}

\newcommand{\xuparrow}[1]{%
  {\left\uparrow\vbox to #1{}\right.\kern-\nulldelimiterspace}
}

\newcommand{\boldparagraph}[1]{\vspace{0.15cm}\noindent{\bf #1.} }

\newcommand{\boldquestion}[1]{\vspace{0.2cm}\noindent{\bf #1} }

\definecolor{myred}{RGB}{178, 34, 34}
\definecolor{mygreen}{RGB}{34, 139, 34}
\definecolor{relacs}{RGB}{184, 84, 80}
\definecolor{internvl}{RGB}{22, 153, 213}
\definecolor{vista}{RGB}{59, 125, 35}

\usepackage[table]{xcolor}
\usepackage{multirow}
\usepackage{makecell}
\usepackage{bbding}

\usepackage[nonatbib,preprint]{neurips_2026}

\usepackage[utf8]{inputenc} % allow utf-8 input
\usepackage[T1]{fontenc}    % use 8-bit T1 fonts
\usepackage{url}            % simple URL typesetting
\usepackage{booktabs}       % professional-quality tables
\usepackage{amsfonts}       % blackboard math symbols
\usepackage{nicefrac}       % compact symbols for 1/2, etc.
\usepackage{microtype}      % microtypography
\usepackage{xcolor}         % colors

\usepackage{multirow}
\usepackage{subcaption}
\usepackage{float}
\usepackage{bbding}

\usepackage{threeparttable}
\usepackage{wrapfig}

\usepackage[pagebackref=true,breaklinks=true,letterpaper=true,colorlinks,bookmarks=false]{hyperref}

\usepackage[capitalize,noabbrev]{cleveref}

\usepackage[numbers,sort&compress]{natbib}

\title{Progressive Risk Estimation for Accident Anticipation}

\author{%
  Samet Hicsonmez\textsuperscript{1,}\thanks{Equal contribution.}
  \quad
  Eray Çakar\textsuperscript{2,}\footnotemark[1]
  \quad
  Nermin Samet\textsuperscript{3,}\thanks{Equal contribution for senior authorship.}
  \quad
  Fatma Güney\textsuperscript{2,}\footnotemark[2]
  \\[0.6em]
  \textsuperscript{1}Independent Researcher, Paris, France\\
  \textsuperscript{2}Department of Computer Engineering, KUIS AI,\\
  Koç University, Istanbul, Turkey\\
  \textsuperscript{3}Valeo.ai, Paris, France
}

\begin{document}

\maketitle

\begin{abstract}
\label{sec:abstract}
% Accident anticipation aims to detect early anomalous driving cues before a crash occurs while avoiding false alarms during normal driving. Current methods typically formulate this task as a binary classification, which overlooks the continuous temporal progression of pre-accident events. We propose \ours to model accident risk as a continuous signal that increases as the accident approaches. To better evaluate global risk behavior, we further introduce a Separation Score that penalizes premature false alarms while rewarding risk increases after anomaly onset. Our method achieves state-of-the-art performance on the CAP and DADA subsets of MM-AU, improving AUC by over 5 points and anticipating accidents up to 0.5 seconds earlier than prior work, while also surpassing the Nexar challenge winner by +1 mAP.
Accident anticipation aims to recognize anomalous driving cues before a crash while avoiding false alarms during normal driving. Existing approaches typically formulate this task as binary classification, focusing on whether an accident will occur rather than \emph{when} it will occur. We propose \ours, a framework that models accident risk as a continuously evolving signal that increases as the crash approaches. By explicitly enforcing temporal ordering and distance-to-accident awareness, our method progressively raises risk while suppressing premature alarms, leading to significant improvements on MM-AU subsets and Nexar. We further introduce a Separation Score to evaluate the global behavior of predicted risk curves beyond local temporal windows. Code and visualizations are available at \url{https://github.com/giddyyupp/PRE-ACT}.
%the CAP and DADA subsets of MM-AU, improving mean AUC by over 10\% and nearly 30\%, respectively, while anticipating accidents up to 0.5 seconds earlier than prior work. It also surpasses the Nexar challenge winner by +1 mAP. 
\end{abstract}

\section{Introduction}
\label{sec:intro}

Despite recent advances in autonomous driving~\cite{jia2025drivetransformer, guo2025ipad,kirby2026driving}, deploying these systems reliably in the real world remains challenging due to the lack of preventive mechanisms in safety-critical scenarios~\cite{nayal2023rba,vojivr2024pixood,galesso2024diffusion,shoeb2025out,gerstenecker2026fail2drive}. In this paper, we take a safety-first approach and propose a method to anticipate accidents early enough to prevent them. Given videos capturing the moments leading up to an accident, our goal is to develop an online monitoring system that recognizes early anomalous cues and issues warnings at each time step. Beyond timely and accurate detection of anomalous cues, a key challenge is minimizing false alarms to ensure the system remains reliable and practical for real-world deployment.

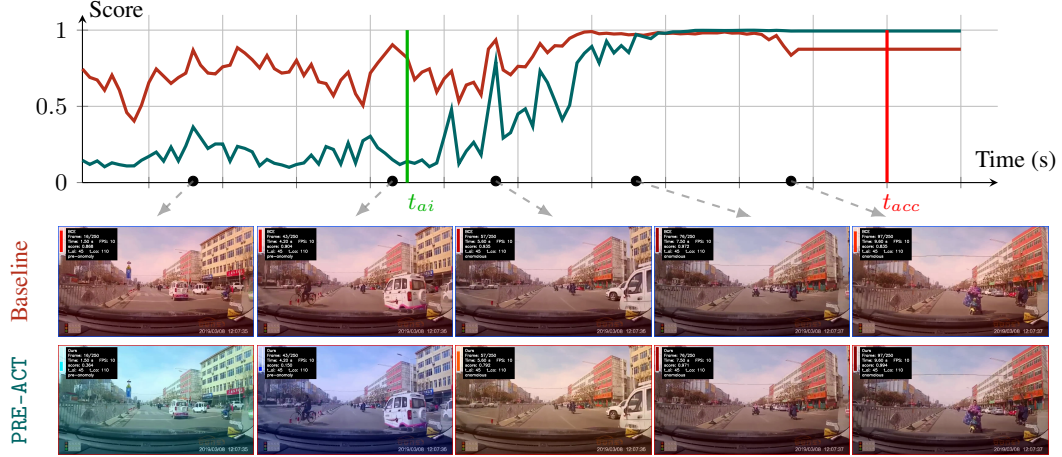
\begin{figure}
\centering
\begin{tikzpicture}

\def\assetdir{figures/teaser_data/5_9511_c26d20ea}
\def\imgw{0.185\textwidth}
\def\xgap{0.188\textwidth}

\def\timeliney{2.2}
\def\bcey{-3.48}
\def\oursy{-5.05}

\node[rotate=90, anchor=center, text=BrickRed, font=\small]
  at (-0.13\textwidth,\bcey) {Baseline};

\node[rotate=90, anchor=center, text=teal!80!black, font=\small]
  at (-0.13\textwidth,\oursy) {\ours};

% BCE row
\node (bce16) at (0*\xgap,\bcey) {\includegraphics[width=\imgw]{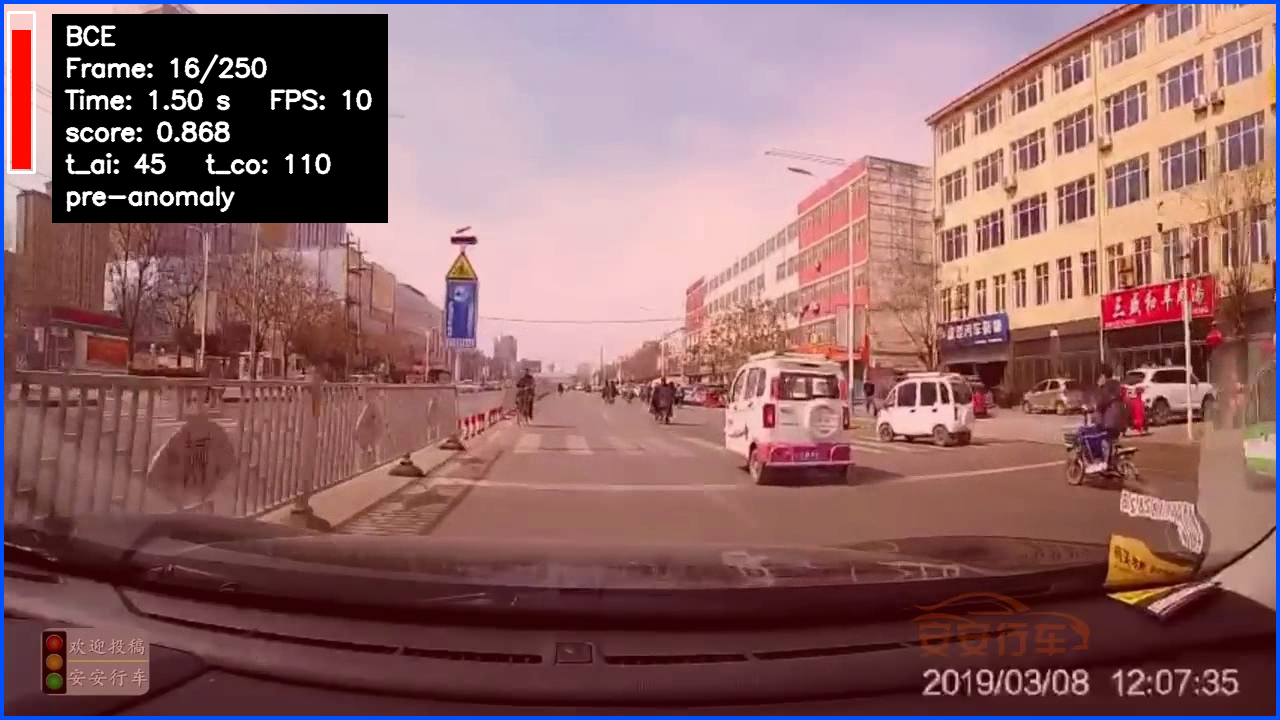}};
\node (bce43) at (1*\xgap,\bcey) {\includegraphics[width=\imgw]{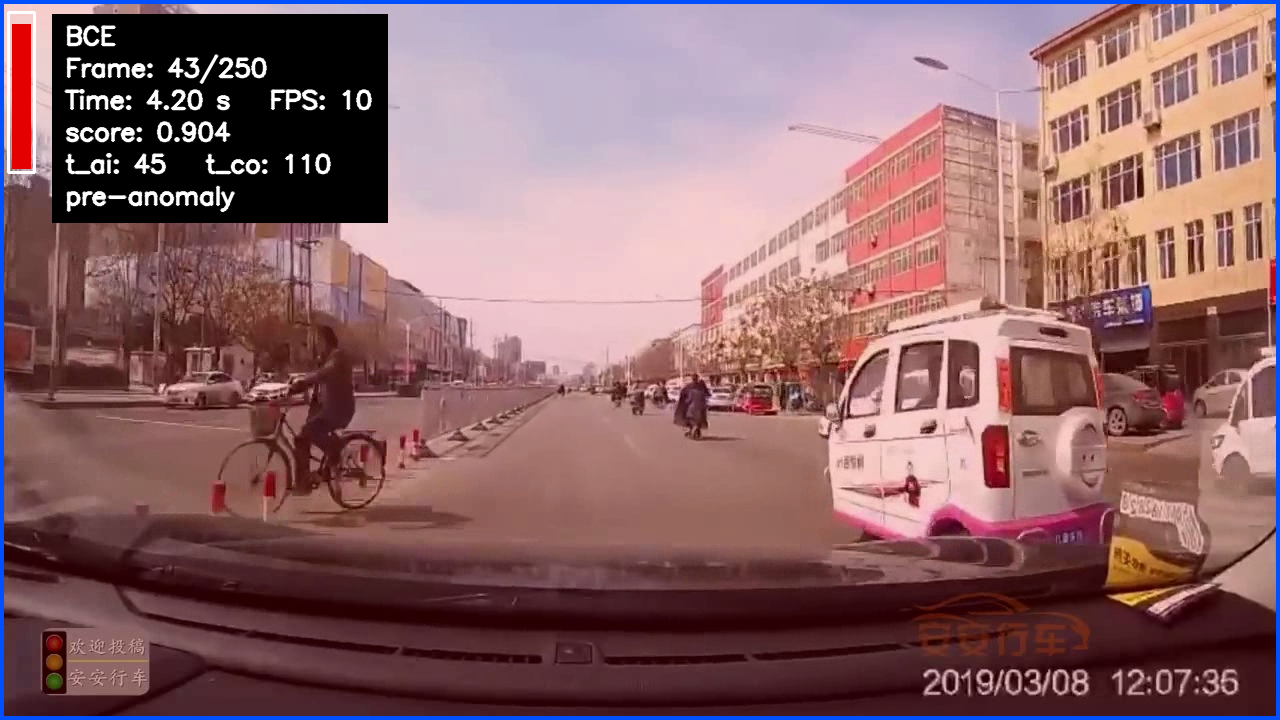}};
\node (bce57) at (2*\xgap,\bcey) {\includegraphics[width=\imgw]{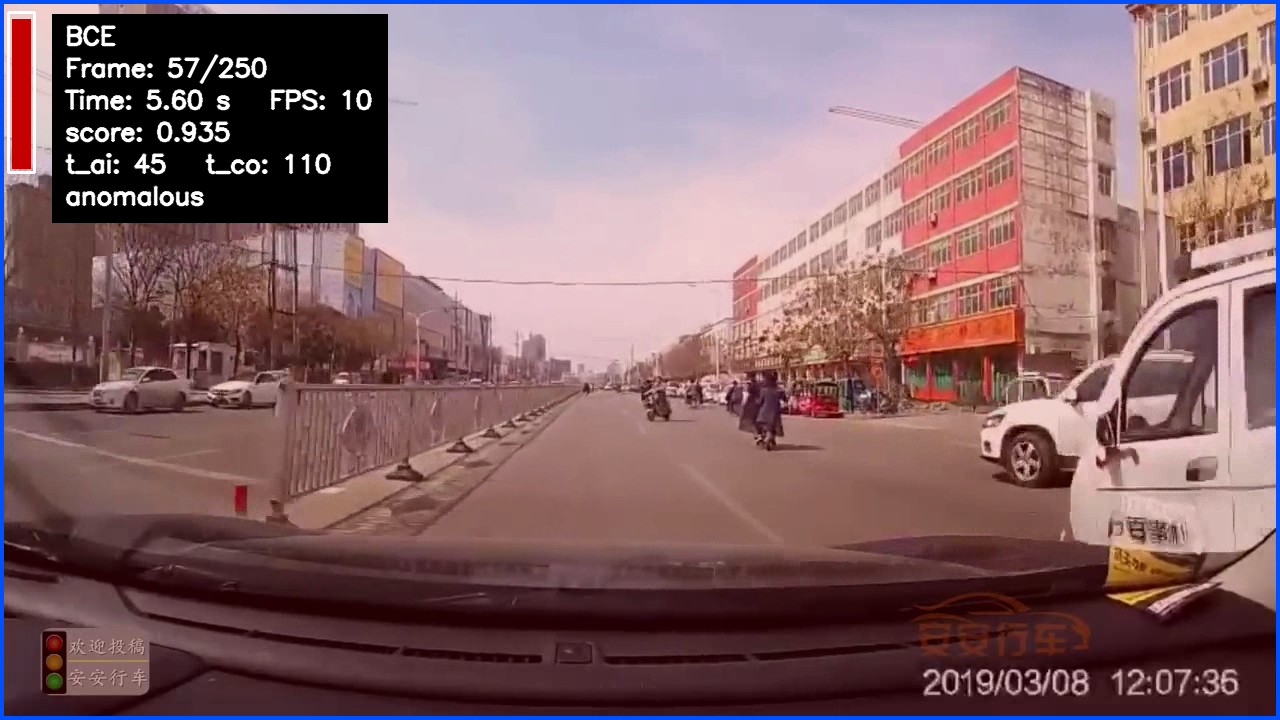}};
\node (bce76) at (3*\xgap,\bcey) {\includegraphics[width=\imgw]{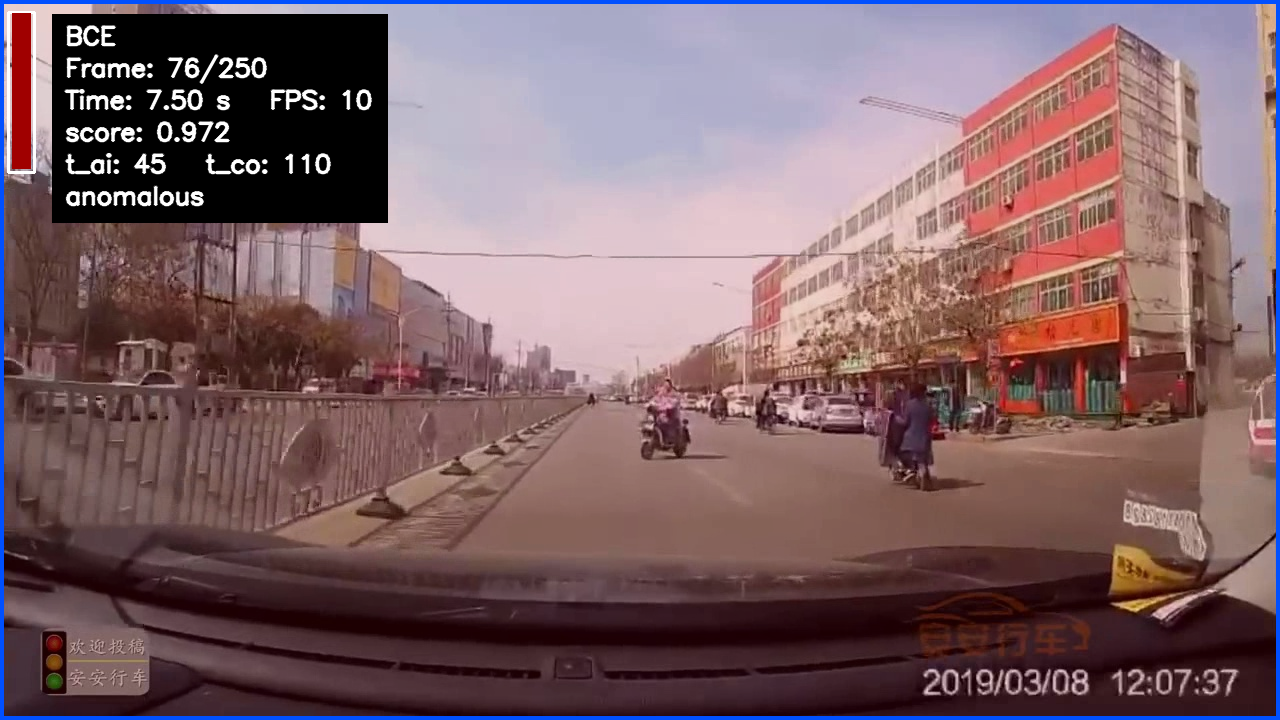}};
\node (bce97) at (4*\xgap,\bcey) {\includegraphics[width=\imgw]{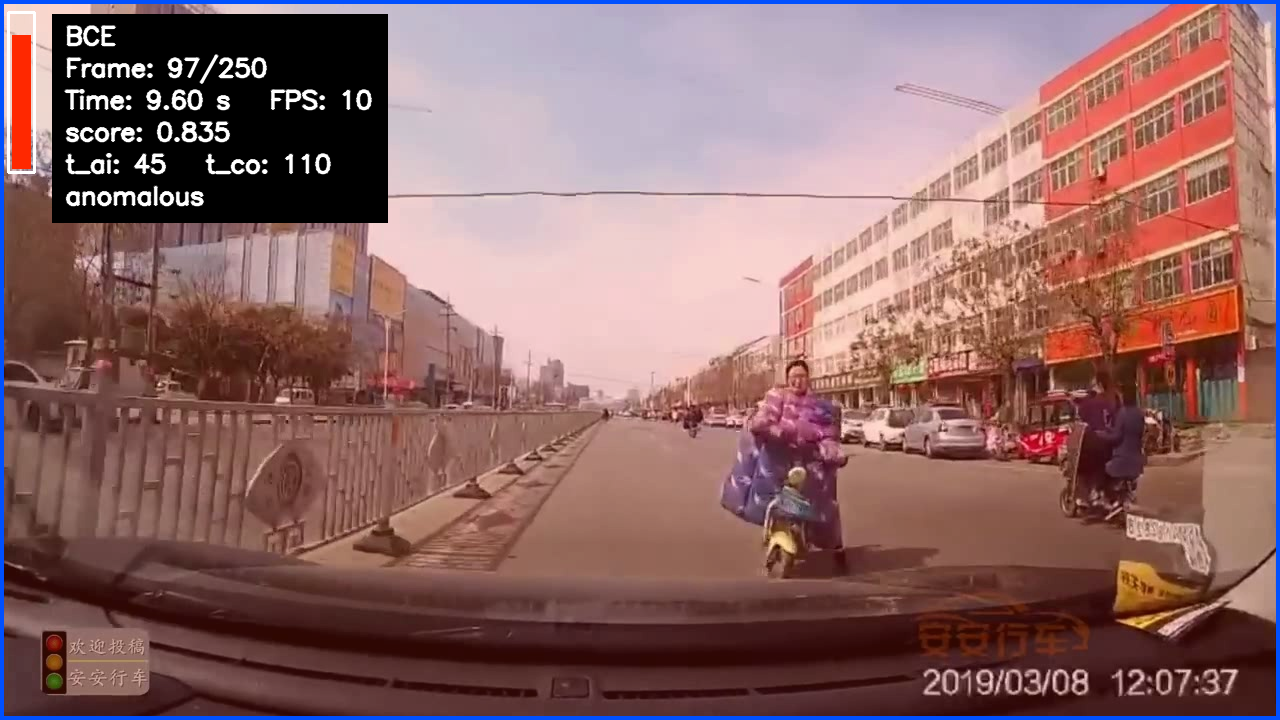}};

% Ours row
\node (ours16) at (0*\xgap,\oursy) {\includegraphics[width=\imgw]{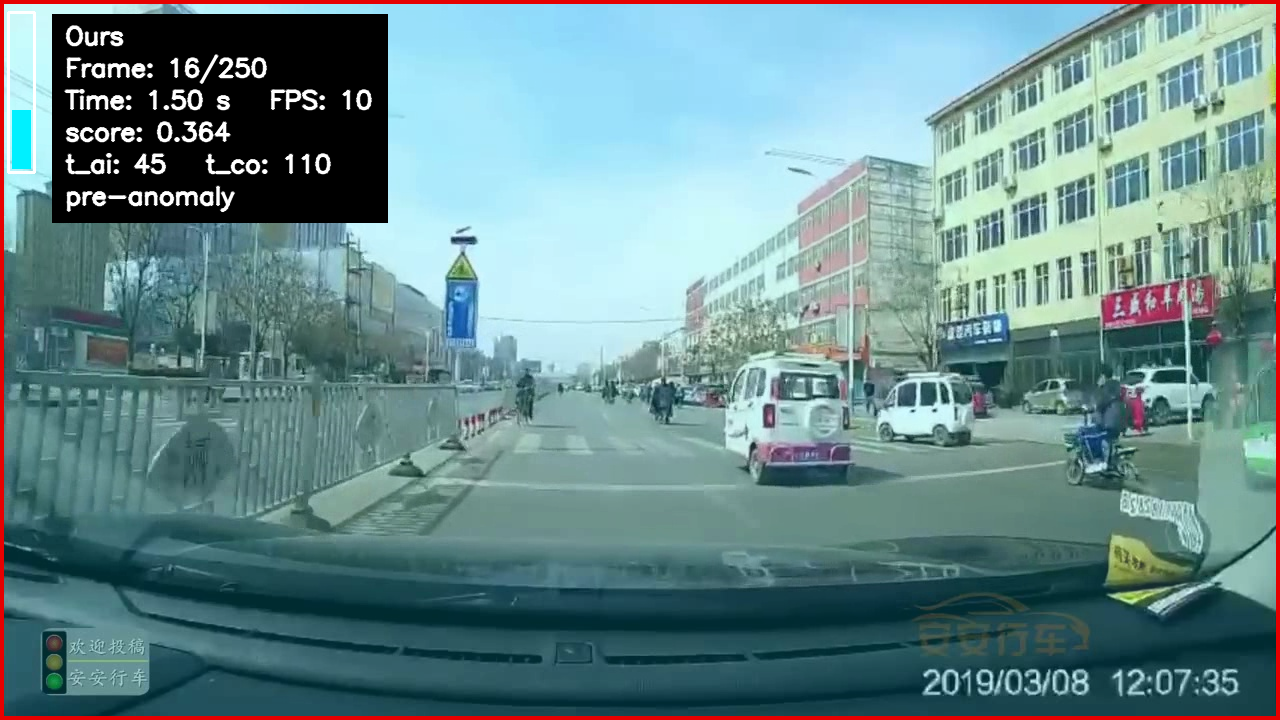}};
\node (ours43) at (1*\xgap,\oursy) {\includegraphics[width=\imgw]{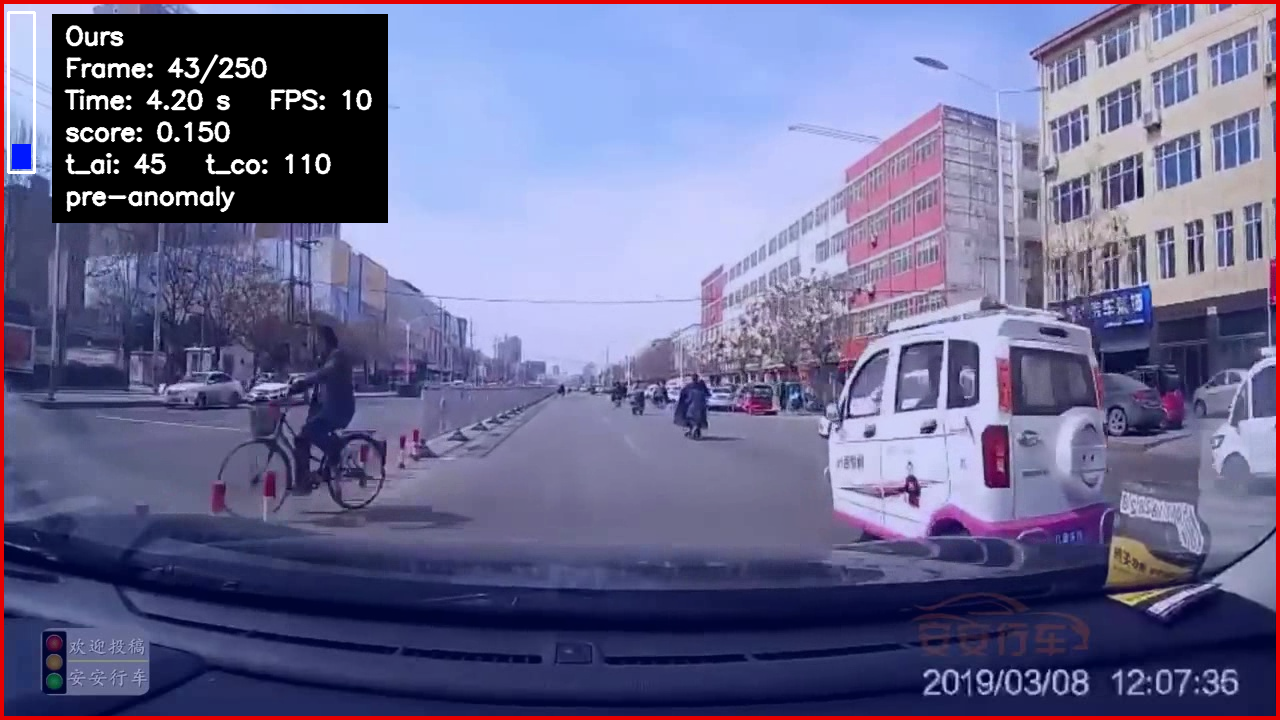}};
\node (ours57) at (2*\xgap,\oursy) {\includegraphics[width=\imgw]{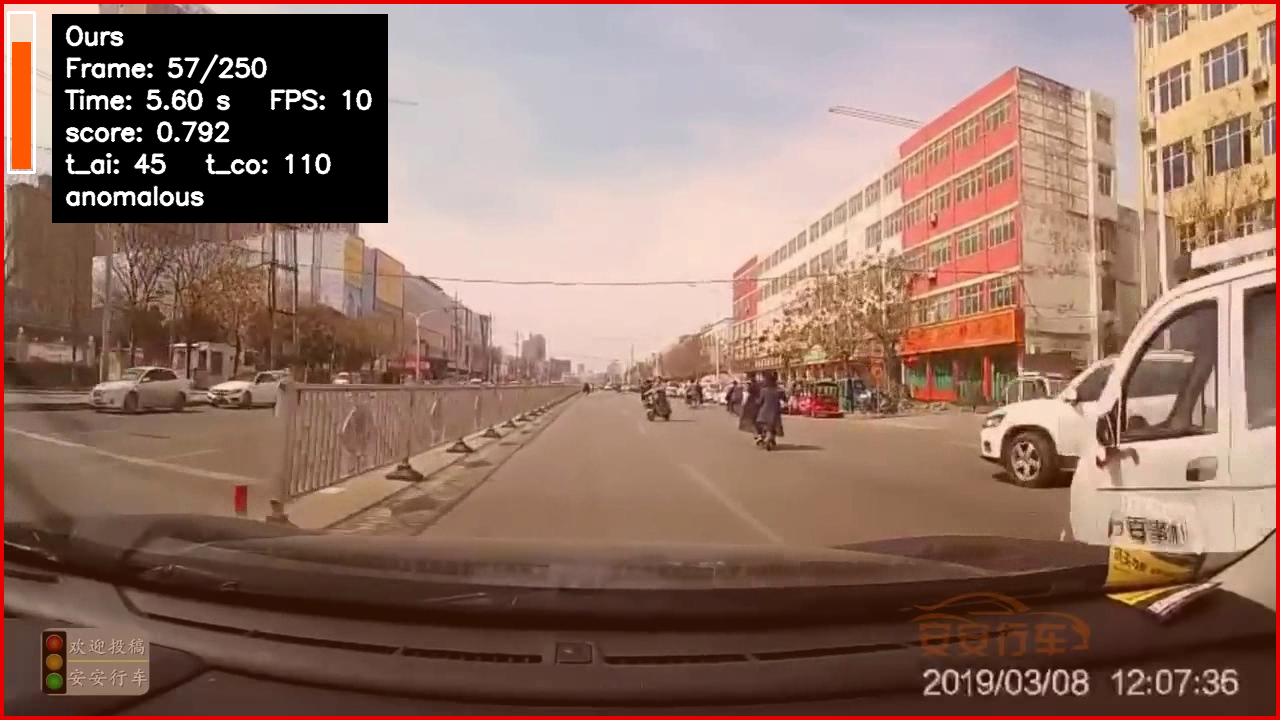}};
\node (ours76) at (3*\xgap,\oursy) {\includegraphics[width=\imgw]{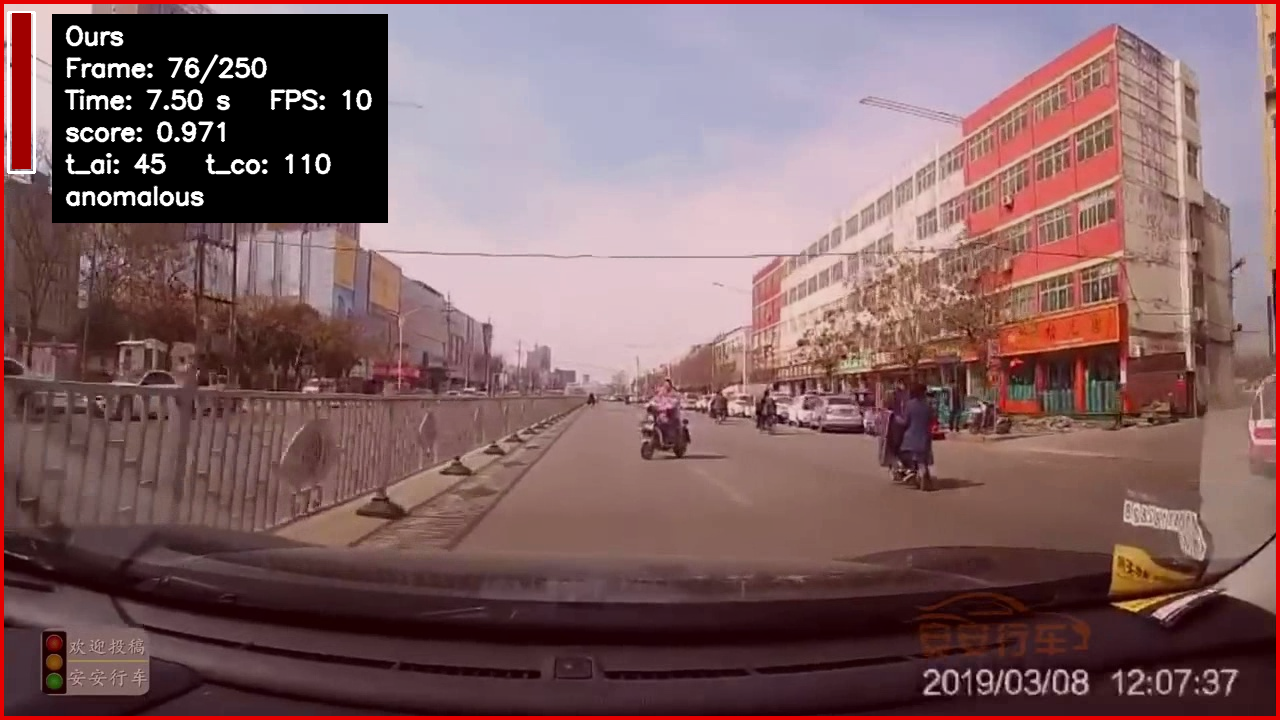}};
\node (ours97) at (4*\xgap,\oursy) {\includegraphics[width=\imgw]{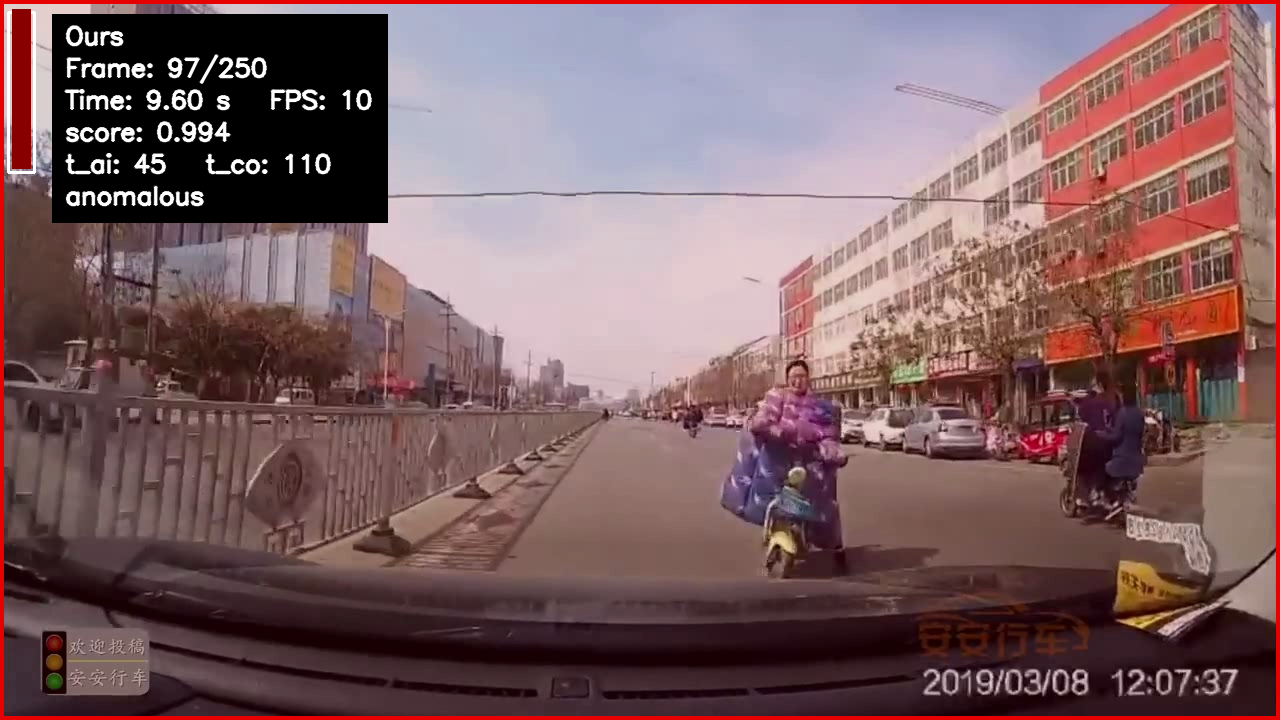}};

% Timeline directly in outer tikzpicture
\begin{axis}[
    at={(-0.07\textwidth,\timeliney)},
    anchor=north west,
    width=0.98\textwidth,
    height=3.8cm,
    xmin=1, xmax=125,
    ymin=0, ymax=1.1,
    ytick={0,0.5,1.0},
    axis lines=left,
    xlabel={},
    ylabel={},
    xticklabels=\empty,
    xlabel style={yshift=-4pt},
    ylabel style={yshift=-3pt},
    tick label style={font=\small},
    grid=major,
    trim axis left,
    trim axis right,
    clip=false,
    axis line style={-{Latex[length=2mm]}},
    axis lines=left,
]

% x-axis label near arrow
\node[anchor=west, font=\small, xshift=2pt, yshift=8pt]
    at (axis cs:120,0) {Time (s)};

% y-axis label near arrow
\node[anchor=south west, font=\small, xshift=-1pt, yshift=-1pt]
    at (axis cs:1,1.04) {Score};

\addplot[BrickRed, very thick, forget plot]
table[x=frame, y=bce, col sep=comma, restrict x to domain=1:120]
{\assetdir/timeline/timeline_bce_ours.csv};

\addplot[teal!80!black, very thick, forget plot]
table[x=frame, y=ours, col sep=comma, restrict x to domain=1:120]
{\assetdir/timeline/timeline_bce_ours.csv};

\addplot[green!70!black, very thick, forget plot]
coordinates {(45,0) (45,1.00)};
\node[green!70!black, font=\bfseries\small, anchor=south]
    at (axis cs:47,-0.25) {$t_{ai}$};

\addplot[red, very thick, forget plot]
coordinates {(110,0) (110,1.00)};
\node[red, font=\bfseries\small, anchor=south]
    at (axis cs:112,-0.25) {$t_{acc}$};

\addplot[only marks, mark=*, mark size=1.8pt, black, forget plot]
coordinates {
    (16,0.010)
    (43,0.010)
    (57,0.010)
    (76,0.010)
    (97,0.010)
};

\coordinate (dot16) at (axis cs:16,0.010);
\coordinate (dot43) at (axis cs:43,0.010);
\coordinate (dot57) at (axis cs:57,0.010);
\coordinate (dot76) at (axis cs:76,0.010);
\coordinate (dot97) at (axis cs:97,0.010);

\end{axis}

% Arrows from black dots to bottom of Ours frames
\draw[dashed, gray, thick, opacity=0.65, -{Latex[length=2mm]}] (dot16) -- (bce16.north);
\draw[dashed, gray, thick, opacity=0.65, -{Latex[length=2mm]}] (dot43) -- (bce43.north);
\draw[dashed, gray, thick, opacity=0.65, -{Latex[length=2mm]}] (dot57) -- (bce57.north);
\draw[dashed, gray, thick, opacity=0.65, -{Latex[length=2mm]}] (dot76) -- (bce76.north);
\draw[dashed, gray, thick, opacity=0.65, -{Latex[length=2mm]}] (dot97) -- ([xshift=-24pt]bce97.north);

\end{tikzpicture}
\caption{\textbf{\ours: From binary accident prediction to progressive risk anticipation.} Unlike the dominant paradigm of formulating accident anticipation as a binary classification of whether an accident will occur (baseline), we model risk as a temporally evolving signal that reflects \emph{when} an accident is likely to occur. By explicitly enforcing temporal order and distance-to-crash awareness, our model reduces false alarms while progressively increasing risk as abnormal cues emerge. Scores are overlaid on frames as an alpha channel, with blue indicating safe and red indicating risky predictions.}
\label{fig:bce_ours_timeline}
\end{figure}

Accidents are inherently dynamic events that unfold over time, requiring models to capture spatio-temporal dependencies and extract meaningful representations from video. Prior work has addressed this by incorporating temporal modeling through recurrent architectures~\cite{karim2022dynamic}, graph-based representations~\cite{wang2023gsc}, or visual attention cues~\cite{zhao2025accident}. However, most existing approaches ultimately formulate accident anticipation as a binary classification problem, typically at the frame level or over a short window~\cite {zhao2025accident}, focusing on whether an accident will occur rather than \emph{when} it will occur (\figref{fig:bce_ours_timeline}).

In contrast, temporal proximity to an accident is a critical signal in anticipation. As an event approaches, subtle cues intensify and evolve, and correctly interpreting this progression requires understanding the temporal ordering and relative distance of frames to the approaching event. Modeling this temporal structure explicitly is therefore essential for timely and reliable accident anticipation. Motivated by this, we propose to model the increasing risk as an accident approaches and introduce a preference over video segments based on their temporal proximity to the event (\figref{fig:model_overview}).

Our method, \ours, advances the state of the art on commonly used accident anticipation datasets, including the CAP and DADA subsets of MM-AU~\cite{fang2024abductive}. It improves the mean AUC by over 10\% on CAP and nearly 30\% on DADA, anticipates crashes 0.25 and 0.18 seconds earlier than prior methods, respectively, and surpasses the winner of the Nexar challenge~\cite{moura2025nexar} by +1 mAP.

Since standard metrics such as AUC, mean time-to-accident (mTTA), and mAP focus on local evaluation windows and may overlook the global behavior of the predicted risk curve, \eg, false alarms during normal driving, we introduce a novel separation score that evaluates global risk behavior by penalizing early false positives and encouraging higher risk after anomaly onset. 

In summary, our contributions are two-fold: \textbf{(i)} We reformulate accident anticipation as explicit continuous risk estimation, leading to significant improvements across all benchmarks and metrics. \textbf{(ii)} We identify a key limitation of existing evaluation metrics and introduce a novel Separation Score to capture the global risk behavior of anticipation models.

% Preamble:
% \usepackage{tikz}
% \usepackage{pgfplots}
% \usetikzlibrary{arrows.meta}
% \pgfplotsset{compat=1.18}
% F8CECC
% B85450
% D5E8D4
% 82B366

\definecolor{digreen}{RGB}{130, 179, 102} % 213, 232, 212	130, 179, 102
\definecolor{dipink}{RGB}{184, 84, 80} % 248, 206, 204 184, 84, 80

%

% Modeling32
% this temporal structure explicitly is therefore essential for timely and reliable accident anticipation.
\section{Related work}

\subsection{Accident anticipation}
\label{subsec:related_anticipation}

Accident anticipation from videos has been framed as an online video classification or anomaly detection problem. Early approaches predominantly relied on RNNs and temporal attention mechanisms to process sequential frames, predicting the likelihood of an accident~\cite{chan2016anticipating, zeng2017agentcentricriskassessmentaccident, Suzuki_2018_CVPR, bao2020uncertainty, fatima2021global, karim2022dynamic, song2024dynamic}.

To capture the dynamic interactions that cause accidents, subsequent research~\cite{mahmood2023new,song2025real,patera2024spatio,li2025dual} has focused heavily on spatiotemporal representation learning for accident anticipation. Graph-based methods, such as GSC~\cite{wang2023gsc} and Graph~\cite{Thakur_2024_WACV}, explicitly model the dynamic interactions between the ego-vehicle and surrounding agents using dynamically updated adjacency matrices. Other works have sought to enrich the input space; for example, by adding 3D information from monocular depth~\cite{liao2024real}. 
Additionally, recent methods CAP~\cite{fang2022cognitive}, CRASH~\cite{liao2024crash}, and Mind-the-Gap~\cite{ryu2026mind} integrate human-inspired cognitive text descriptions or visual attention to boost performance. The underlying learning paradigm in these works generally remains frame-level binary risk scoring.

Recognizing the limitations of static, single-step anomaly scores, recent work has moved toward explicit temporal and future modeling. A notable example is TOP~\cite{zhao2025accident}, which predicts accident probabilities at multiple future timestamps by outputting discrete classification logits for a fixed set of future frames. While this multi-horizon formulation improves anticipation performance and reduces false alarms, it still treats future accident occurrence as a binary target and does not explicitly model the progressive nature of risk. In contrast, we share TOP's motivation of anticipating future risk, but adopt a more structured formulation by changing the supervisory signal to a monotonically increasing risk function that captures how accident risk evolves as the crash approaches.

\subsection{Progress estimation}

Prior works initially captured task progress through embedding-based visual representations~\cite{ma2023vipuniversalvisualreward, ma2023livlanguageimagerepresentationsrewards, nair2022r3muniversalvisualrepresentation}
or via VQA-style binary success classifiers~\cite{du2023visionlanguagemodelssuccessdetectors}. 
More recently, the focus has shifted toward foundation models to build generalist progress estimators. Recent work has explored fine-tuning VLMs on large-scale datasets of successful and failed trajectories to predict discretized progress scores~\cite{lee2026roborewardgeneralpurposevisionlanguagereward}, as well as modeling distance-to-goal values for precise robotic manipulation~\cite{tan2025robodopaminegeneralprocessreward}. To improve generalization without relying heavily on human preference annotations, other approaches introduce auxiliary preference prediction objectives that compare heterogeneous trajectories alongside direct progress estimation~\cite{liang2026robometerscalinggeneralpurposerobotic}. Concurrently, a training-free line of research investigates zero-shot VLMs as temporal value estimators using in-context trajectory shuffling~\cite{budzianowski2026opengvlbenchmarkingvisual,ma2024visionlanguagemodelsincontext}, while alternative methods extract continuous progress signals directly from internal token probabilities to avoid the instability of numeric VLM outputs~\cite{chen2026toprewardtokenprobabilitieshidden}.

Inspired by these dense temporal formulations, our work adapts progress estimation to the domain of accident anticipation by inverting the concept: rather than tracking the closeness to task success, we track the temporal proximity to catastrophic failure. By modeling accident anticipation as a monotonically increasing ``risk progression'', we provide our system with the dense temporal supervisory signal necessary to learn the escalating danger in driving videos.
\section{Method}
\label{sec:method}
%
% Preamble:
% \usepackage{tikz}
% \usetikzlibrary{arrows.meta,decorations.pathreplacing,calc}

\def\framew{1.5cm}
\def\frameh{1.0cm} % 2.1 * 720 / 1280
\definecolor{digreen}{RGB}{225, 213, 231} % 213, 232, 212	130, 179, 102
\definecolor{dipink}{RGB}{248, 206, 204} % 248, 206, 204 184, 84, 80

\def\framepos{1.5}

\begin{figure}[t]
\centering
\begin{tikzpicture}[
    font=\small,
    frame/.style={draw, thick, minimum width=0.72cm, minimum height=0.72cm},
    backbone/.style={draw, thick, trapezium, trapezium left angle=75, trapezium right angle=75,
        minimum width=2.2cm, minimum height=1.45cm, align=center},
head/.style={
    draw,
    thick,
    rounded corners=2pt,
    minimum width=1.7cm,
    minimum height=0.9cm,
    align=center,
    fill=orange!10
},
    arrow/.style={-{Latex[length=2.2mm]}, thick},
    redarrow/.style={-{Latex[length=2.2mm]}, very thick, red!75!black},
    greenarrow/.style={-{Latex[length=2.2mm]}, very thick, teal!80!black},
]

% ---------------- timeline ----------------
\draw[-{Latex[length=4mm]}, very thick] (-6.8,3.0) -- (6.8,3.0)
    node[above left=2pt] {Time};

% Time axis additions
\coordinate (timeStart) at (-6.8,3.0);
\coordinate (timeEnd)   at (6.8,3.0);
\coordinate (thiMark)   at (-2.25,3.0);
% \coordinate (tiMark)   at ($(c4.south east |- timeStart)$);
\coordinate (tcoMark)   at (5.60,3.0);

% \draw[thick] (tiMark) -- ++(0,-0.12);
% \node[below=2pt] at (tiMark) {$t_i$};

\draw[thick] (thiMark) -- ++(0,0.15);
\node[above=2pt] at (thiMark) {$t_h$};

\draw[thick] (tcoMark) -- ++(0,0.15);
\node[above=2pt] at (tcoMark) {$t_{acc}$};

% BCE critical zone -> risk scores
\shade[left color=red!10, right color=red!100, opacity=0.85]
  ([yshift=-0.25cm]thiMark) rectangle ([yshift=-0.65cm]tcoMark);
\node[font=\scriptsize, anchor=south] 
  at ($(thiMark)!0.5!(tcoMark)+(0,-0.65)$) {Critical Zone};

% BCE safe zone -> 0s
\shade[left color=green!80, right color=green!80, opacity=0.65]
  ([yshift=-0.25cm]timeStart) rectangle ([yshift=-0.65cm]thiMark);
\node[font=\scriptsize, anchor=south] 
  at ($(timeStart)!0.5!(thiMark)+(0,-0.65)$) {Safe Zone};

% heatmap critical zone -> risk scores
% \shade[left color=green!80, right color=red!80, opacity=0.85]
%   ([yshift=-0.80cm]thiMark) rectangle ([yshift=-1.20cm]tcoMark);
% \node[font=\scriptsize, anchor=south] 
%   at ($(thiMark)!0.5!(tcoMark)+(0,-1.25)$) {increasing risk};

% heatmap safe zone -> 0s
% \shade[left color=Goldenrod!20, right color=Goldenrod!20, opacity=0.85]
%   ([yshift=-0.80cm]timeStart) rectangle ([yshift=-1.20cm]thiMark);
% \node[font=\scriptsize, anchor=south] 
%   at ($(timeStart)!0.5!(thiMark)+(0,-1.20)$) {No Risk};

% % ---------------- Initial frames ----------------
\node[inner sep=0pt, draw, thick] (c10) at (-6.0,\framepos) {\includegraphics[width=\framew,height=\frameh]{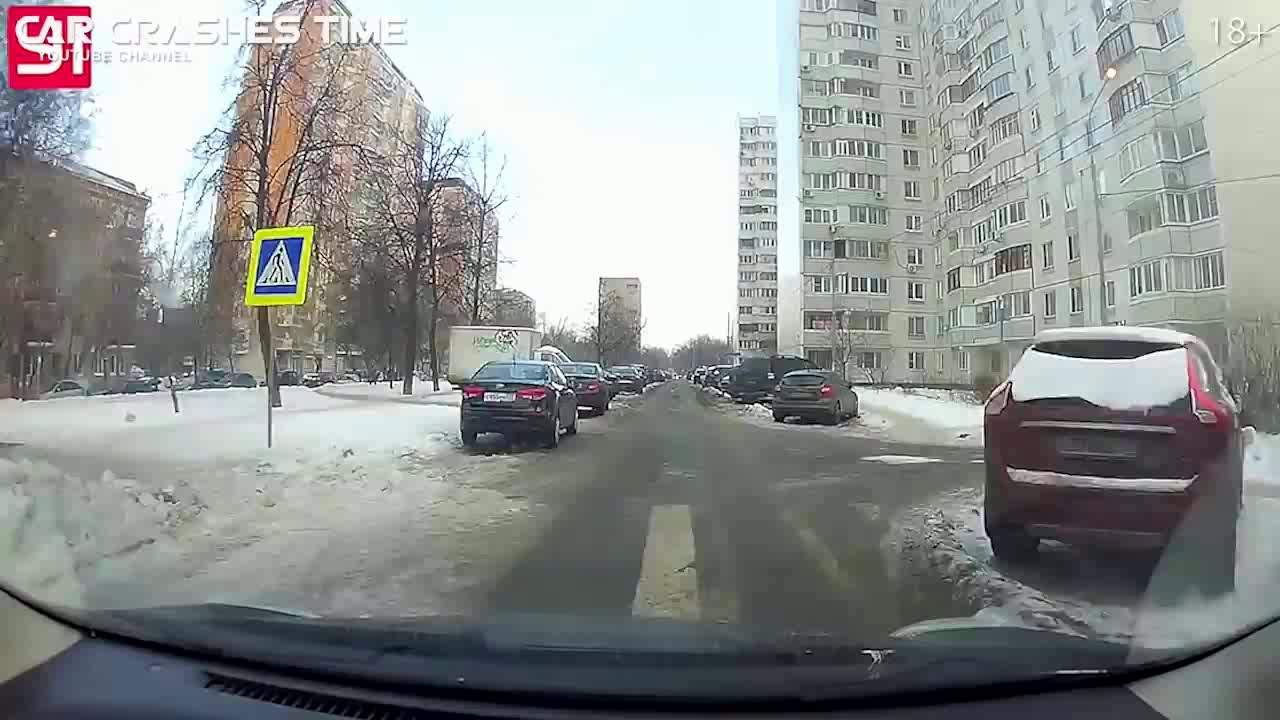}};

\node at (-5.0,\framepos) {$\cdots$};

\node[inner sep=0pt, draw, thick] (c14) at (-4.0,\framepos) {\includegraphics[width=\framew,height=\frameh]{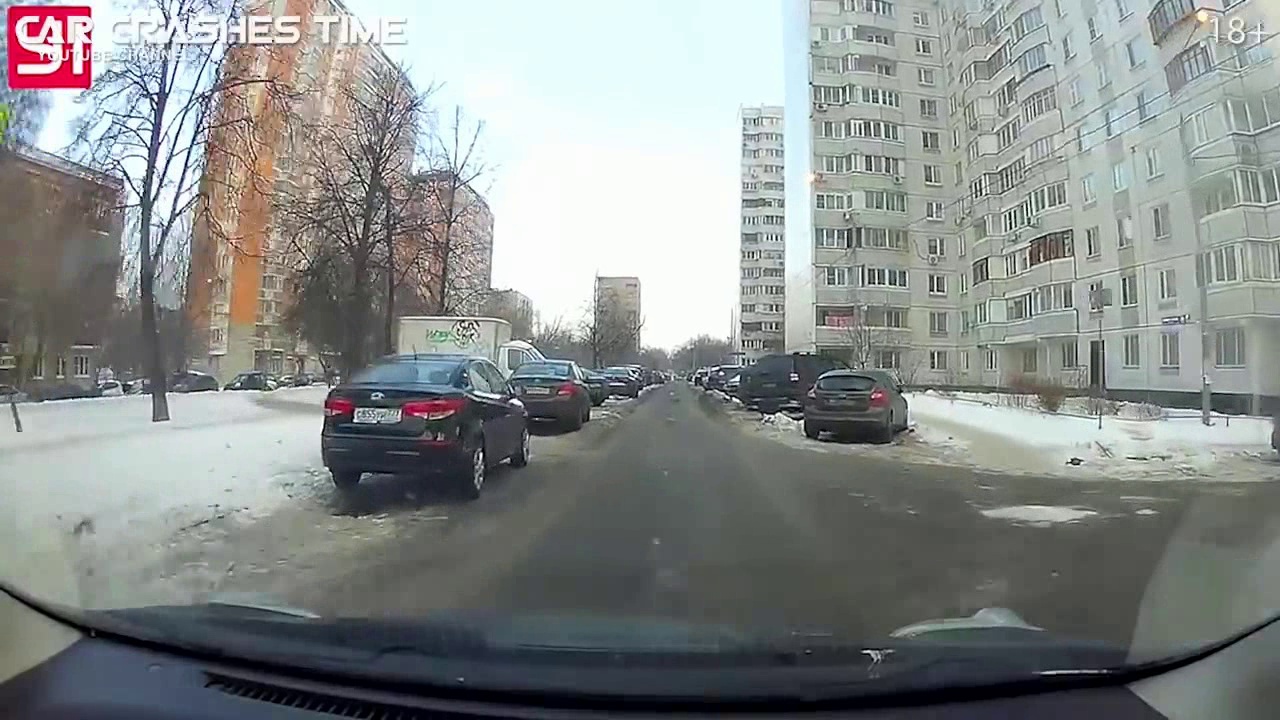}};

\coordinate (c1BraceLeft)  at ($(c10.south west)+(0,-0.3)$);
\coordinate (c1BraceRight) at ($(c14.south east)+(0,-0.3)$);
\coordinate (c1brace) at ($(c1BraceLeft)!0.5!(c1BraceRight)$);

% brace over current N frames
\draw[decorate, decoration={brace, mirror, amplitude=5pt}]
    ($(c10.south west)+(0,-0.08)$) -- ($(c14.south east)+(0,-0.08)$)
    coordinate[midway] (c1brace);

% Label
\node[fill=white, inner sep=1pt] at ($(c1brace)+(0,-0.45)$)
  {{clip}$_1$};

\node at (-2.75,\framepos) {$\cdots$$\cdots$};

% MIDDLE FRAMES
\node[inner sep=0pt, draw, thick] (c20) at (-1.5,\framepos) {\includegraphics[width=\framew,height=\frameh]{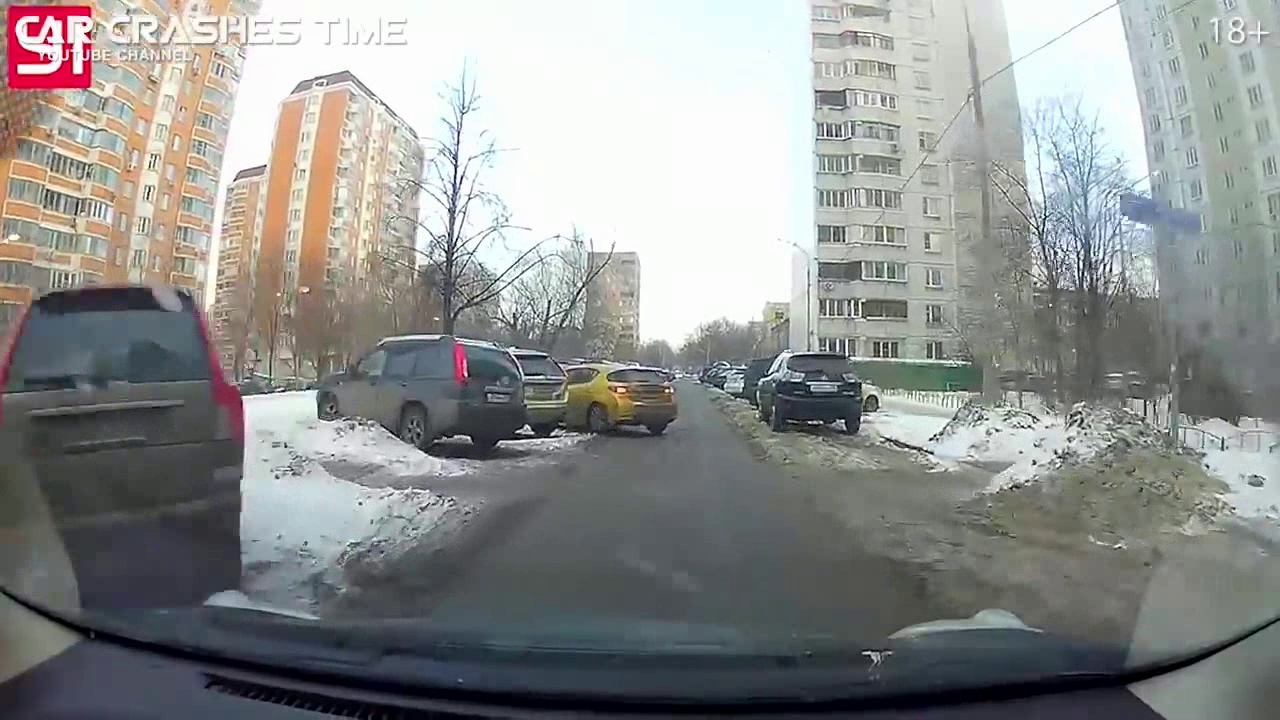}};

\node at (-0.5,\framepos) {$\cdots$};

\node[inner sep=0pt, draw, thick] (c24) at (0.5,\framepos) {\includegraphics[width=\framew,height=\frameh]{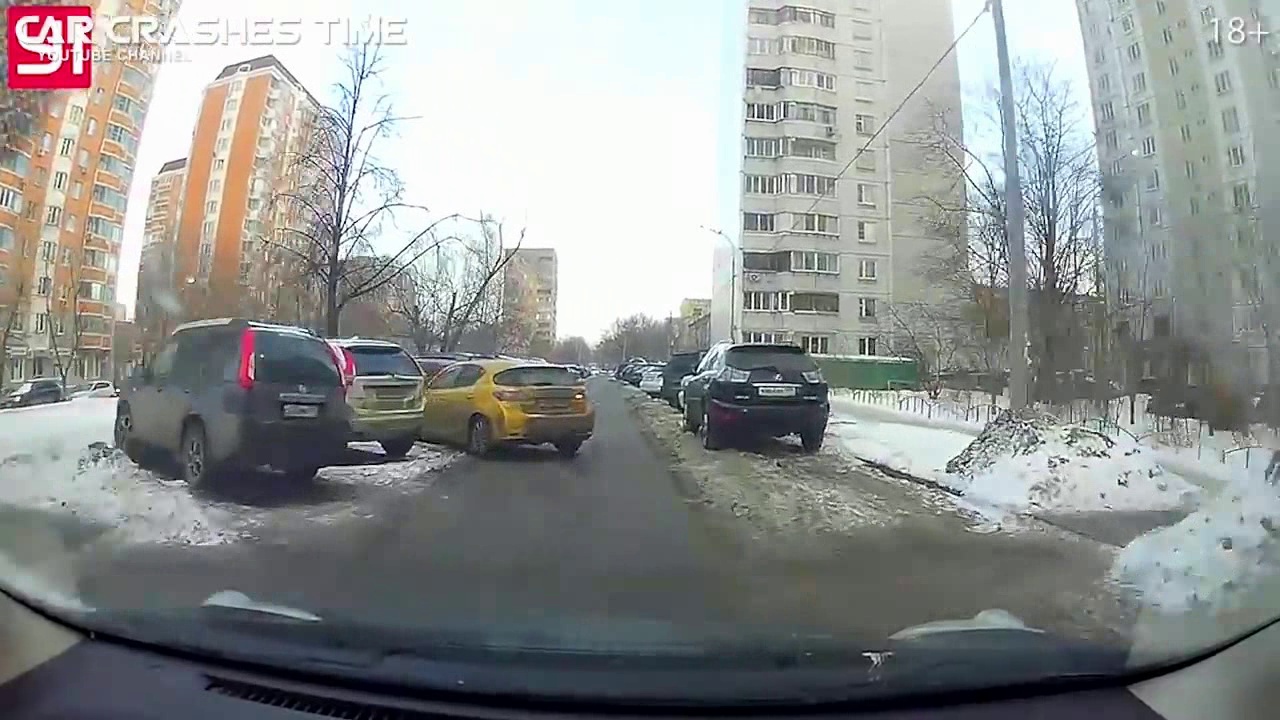}};

\coordinate (c2BraceLeft)  at ($(c20.south west)+(0,-0.3)$);
\coordinate (c2BraceRight) at ($(c24.south east)+(0,-0.3)$);
\coordinate (c2brace) at ($(c2BraceLeft)!0.5!(c2BraceRight)$);

% brace over current N frames
\draw[decorate, decoration={brace, mirror, amplitude=5pt}]
    ($(c20.south west)+(0,-0.08)$) -- ($(c24.south east)+(0,-0.08)$)
    coordinate[midway] (c2brace);

% Label
\node[fill=white, inner sep=1pt] at ($(c2brace)+(0,-0.45)$)
  {{clip}$_i$};

\node at (1.72,\framepos) {$\cdots$$\cdots$};

% ---------------- end frames ----------------
\node[inner sep=0pt, draw, thick] (c30) at (2.85,\framepos) {\includegraphics[width=\framew,height=\frameh]{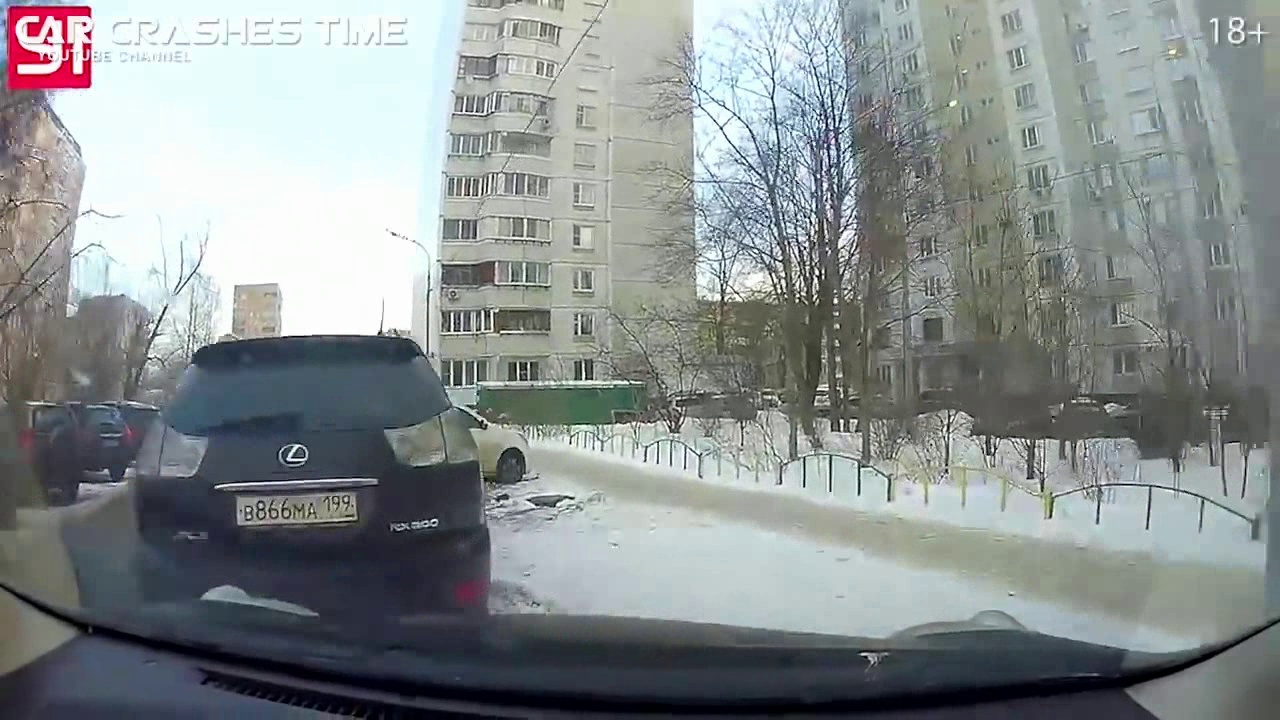}};

\node at (3.85,\framepos) {$\cdots$};

\node[inner sep=0pt, draw, thick] (c34) at (4.85,\framepos) {\includegraphics[width=\framew,height=\frameh]{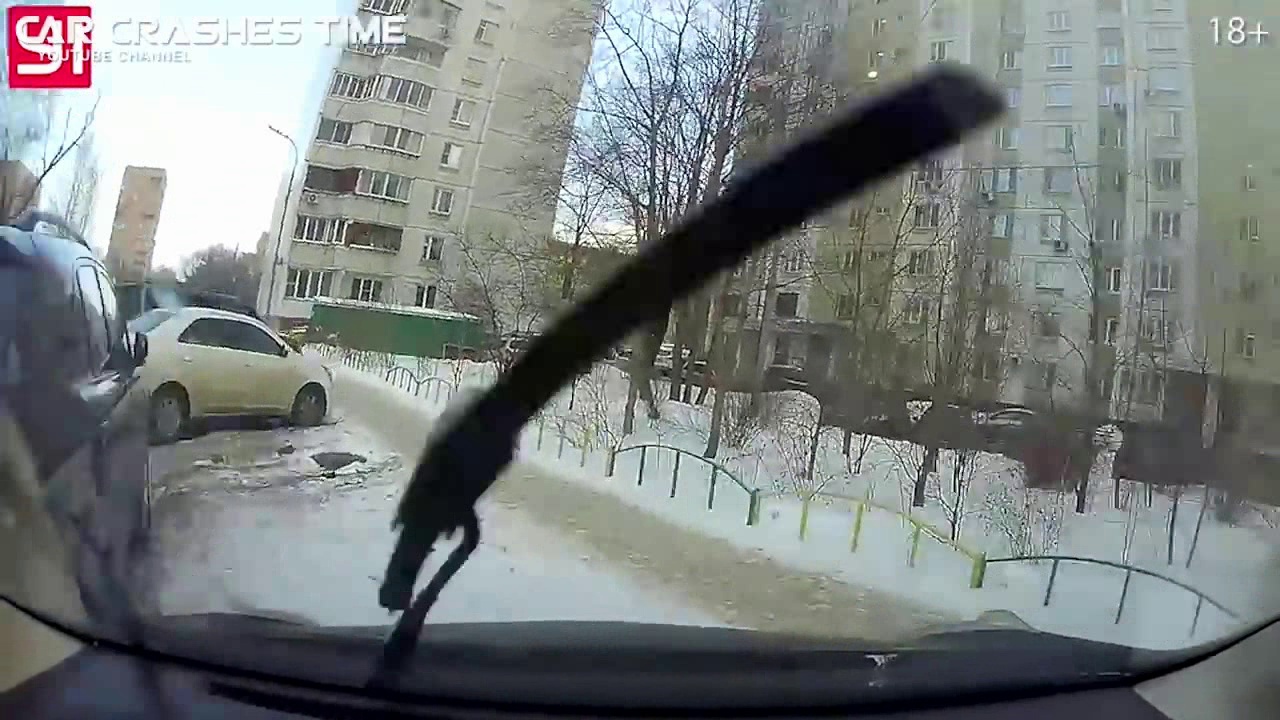}};

\coordinate (c3BraceLeft)  at ($(c30.south west)+(0,-0.3)$);
\coordinate (c3BraceRight) at ($(c34.south east)+(0,-0.3)$);
\coordinate (c3brace) at ($(c3BraceLeft)!0.5!(c3BraceRight)$);

% brace over current N frames
\draw[decorate, decoration={brace, mirror, amplitude=5pt}]
    ($(c30.south west)+(0,-0.08)$) -- ($(c34.south east)+(0,-0.08)$)
    coordinate[midway] (c3brace);

% Label
\node[fill=white, inner sep=1pt] at ($(c3brace)+(0,-0.45)$)
  {{clip}$_j$};

% =========================================================
% Separation line
% =========================================================

% \draw[thick] (-6.8,-0.4) -- (6.2,-0.4);

% ---------------- stacked N frames: bottom-left ----------------

% Main visible frame
\node[inner sep=0pt, draw, thick] (s0) at (-6.0,-1.00)
    {\includegraphics[width=\framew,height=\frameh]{figures/model_images/000051.jpg}};

% Back frames, slightly shifted
\node[inner sep=0pt, draw, thick] (s1) at ($(s0)+(0.16,0.12)$)
    {\includegraphics[width=\framew,height=\frameh]{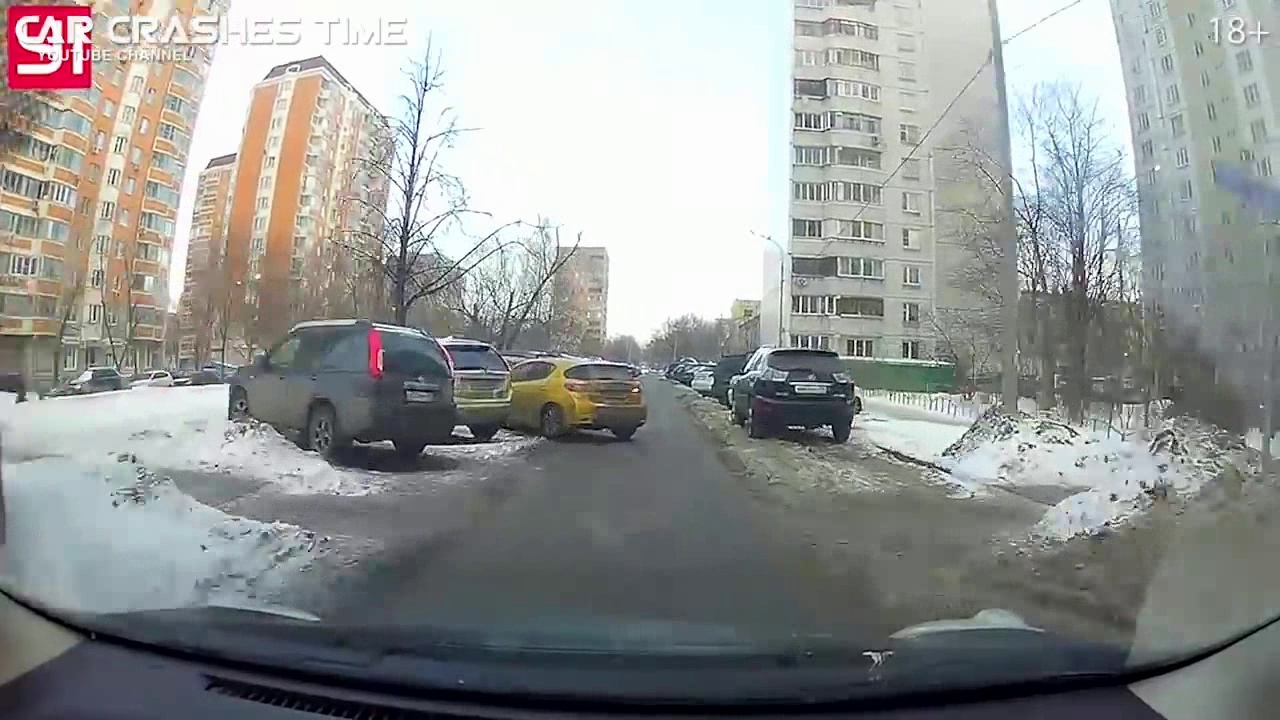}};

\node[inner sep=0pt, draw, thick] (s2) at ($(s0)+(0.32,0.24)$)
    {\includegraphics[width=\framew,height=\frameh]{figures/model_images/000055.jpg}};

% Clip label
\node[below=0.18cm of s2] {clip$_i$};

% Main visible frame
\node[inner sep=0pt, draw, thick] (t0) at (-6.0,-3.0)
    {\includegraphics[width=\framew,height=\frameh]{figures/model_images/000066.jpg}};

% Back frames, slightly shifted
\node[inner sep=0pt, draw, thick] (t1) at ($(t0)+(0.16,0.12)$)
    {\includegraphics[width=\framew,height=\frameh]{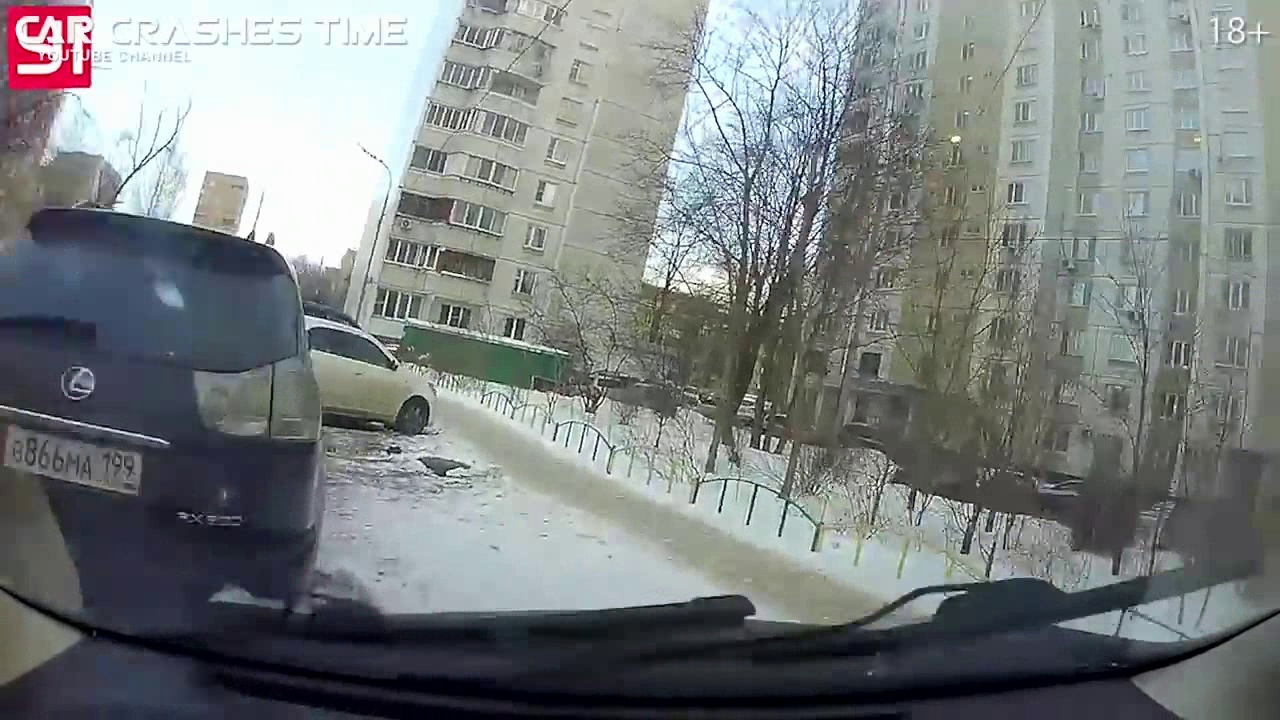}};

\node[inner sep=0pt, draw, thick] (t2) at ($(t0)+(0.32,0.24)$)
    {\includegraphics[width=\framew,height=\frameh]{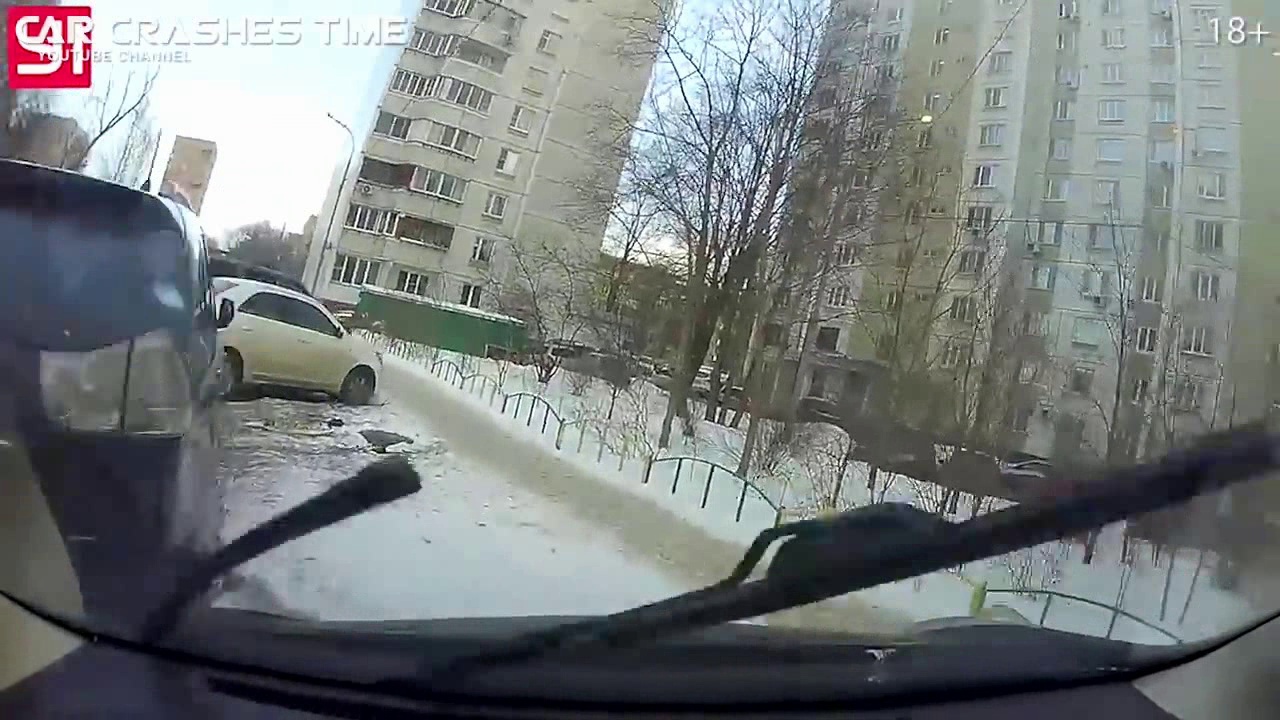}};

% Clip label
\node[below=0.18cm of t2] {clip$_j$};

% ---------------- bottom horizontal model pipeline ----------------

% Video Foundation Model
\node[
    draw,
    thick,
    rounded corners=3pt,
    minimum width=2.4cm,
    minimum height=1.25cm,
    align=center,
    fill=SpringGreen!10
    % fill=Turquoise!20
] (fm) at (-2.9,-1.8) {Video\\Foundation\\Model};

% Heads
\node[
    draw,
    thick,
    rounded corners=3pt,
    minimum width=1.65cm,
    minimum height=0.65cm,
    align=center,
    fill=orange!10
] (bce) at (0.0,-0.80) {Accident \\ Head};

\node[
    draw,
    thick,
    rounded corners=3pt,
    minimum width=1.65cm,
    minimum height=0.65cm,
    align=center,
    fill=orange!10
] (prog) at (0.0,-1.80) {Progress \\ Head};

\node[
    draw,
    thick,
    % dashed,
    rounded corners=3pt,
    minimum width=1.65cm,
    minimum height=0.65cm,
    align=center,
    fill=orange!10
] (pref) at (0.0,-2.80) {Preference \\ Head};

% Inputs from stacked clips
\draw[-{Latex[length=2.2mm]}, thick]
    (s2.east) -- ($(fm.west)+(0,0.32)$);

\draw[-{Latex[length=2.2mm]}, thick, dashed]
    (t2.east) -- ($(fm.west)+(0,-0.32)$);

% FM to heads
\draw[-{Latex[length=2.2mm]}, thick]
    (fm.east) -- (bce.west);

\draw[-{Latex[length=2.2mm]}, thick]
    (fm.east) -- (prog.west);

\draw[-{Latex[length=2.2mm]}, thick, dashed]
    (fm.east) -- (pref.west);

% Output labels
% \node[right=0.35cm of bce] {safe / critical?};
% \node[right=0.35cm of prog] {risk score};
\node[right=0.15cm of pref] {$j > i$ ?};

% Safe / critical split label
\node[right=0.15cm of bce, inner sep=0pt] (safecrit) {
\begin{tikzpicture}[baseline=(current bounding box.center)]
    \node[
        draw=green!80,
        fill=green!80,
        fill opacity=0.65,
        draw opacity=0.65,
        text opacity=1,
        rounded corners=0pt,
        minimum height=0.55cm,
        minimum width=1.0cm
    ] (safe) at (0,0) {safe};

    \node[
        right=1pt of safe,
        inner sep=1pt
    ] (slash) {/};

    \node[
        draw=red!100,
        fill=red!100,
        fill opacity=0.65,
        draw opacity=0.65,
        text opacity=1,
        rounded corners=0pt,
        minimum height=0.55cm,
        minimum width=1.2cm,
        right=1pt of slash
    ] (critical) {critical};

    \node[
        right=1pt of critical,
        inner sep=1pt
    ] (qmark) {?};
\end{tikzpicture}
};

% Risk score heatmap label
\node[right=0.15cm of prog, inner sep=0pt] (riskmap) {
\begin{tikzpicture}[baseline=(current bounding box.center)]
    \shade[left color=green!80, right color=red!90]
        (0,0) rectangle (2.45,0.55);

    % \draw[thick, rounded corners=2pt]
    %     (0,0) rectangle (2.2,0.55);

    \node (risk) at (0.6,0.275) {risk score};

    \node[
        right=19pt of risk,
        inner sep=1pt
    ] (qmark) {?};
\end{tikzpicture}
};

% Head architecture inset
\node[draw, rounded corners=2pt, fill=orange!10, inner sep=3pt] (headarch) at (5.7,-1.8) {
\begin{tikzpicture}[
    layer/.style={draw, rounded corners=1pt, minimum width=1.8cm,
                  minimum height=0.35cm, font=\scriptsize, fill=white},
    arr/.style={-{Latex[length=1.5mm]}, thin}
]
\node[layer] (l1) {LayerNorm};
\node[layer, below=0.08cm of l1] (l2) {Linear $d{\to}d$};
\node[layer, below=0.08cm of l2] (l3) {GELU};
\node[layer, below=0.08cm of l3] (l4) {Dropout};
\node[layer, below=0.08cm of l4] (l5) {Linear $d{\to}1$};

\draw[arr] (l1) -- (l2);
\draw[arr] (l2) -- (l3);
\draw[arr] (l3) -- (l4);
\draw[arr] (l4) -- (l5);
\end{tikzpicture}
};

\node[
    below=5pt of headarch,
    inner sep=1pt
] (headinfo) {\shortstack{Head  Architecture}};

\end{tikzpicture}
\caption{\textbf{Overview}.
We define the period before the fixed anticipation horizon $t_h$ as the safe zone, and the interval from $t_h$ to the accident time $t_{\text{acc}}$ as the critical zone with progressively increasing risk. During training, we extract spatio-temporal features using a video foundation model. Based on these features, the model classifies clips as safe or critical, assigns a continuous risk score, and ranks them using preference head outputs, encouraging higher scores for critical clips closer to the accident. Dashed arrows indicate components used only during training.
%We define the period before the fixed anticipation horizon $t_h$ as the safe zone, while the interval from $t_h$ to the accident time $t_{acc}$ is treated as the critical zone with progressively increasing risk. During training, we first extract spatio-temporal features using a video foundation model. Based on these features, the model determines whether each clip belongs to the safe or critical zone, assigns a continuous risk score, and ranks clips using the output of the preference head scores $\hat{p}$, encouraging higher scores for clips closer to the accident. Dashed arrows indicate components used only during training.
}
\label{fig:model_overview}
% \vspace{-3mm}
\end{figure}

In this paper, we develop a video monitoring system to anticipate accidents before they happen, \ie, in an online manner. Given the driving situation observed from the perspective of the ego-vehicle, our goal is to continuously increase the warning signal if the vehicle is on a trajectory likely to result in an accident. We hypothesize that there are observable cues in driving videos that intensify as an accident approaches; for example, a vehicle appearing increasingly larger before impact. To capture such cues, we first divide videos into temporal zones based on their proximity to an accident (\secref{subsec:problem_formulation}).
Then, we approach this problem as a video processing task and process each video relying on progress in video foundation models (\secref{subsec:architecture}).
Finally, we propose a novel way of estimating the continuously increasing risk of an accident (\secref{subsec:progress_estimation}).
See \figref{fig:model_overview} for an overview.

\subsection{Problem formulation}
\label{subsec:problem_formulation}

We process driving videos online without assuming access to future information. At any time $t \in \{1, \dots, T\}$ in a video of length $T$, we maintain a state $\bs_t$ that is evaluated in terms of its likelihood of leading to an accident $a_t \in [0,1]$, with 1 representing moments that lead to an accident. 

Accident anticipation datasets provide the time of the accident $t_{\text{acc}}$, but only a subset of them includes an anomaly onset label $t_{ai}$, denoting the first time step at which accident-related cues become visible. While this label is commonly used for evaluation, recent methods~\cite{zhao2025accident} avoid using it during training to reduce supervision. Moreover, such annotations are often subjective, dataset-dependent, and not consistently available across benchmarks; \eg, they are not provided on Nexar~\cite{moura2025nexar}. 
We, therefore, use a fixed anticipation horizon $t_h$ to partition each sequence into two temporal zones:
\begin{itemize}
    \item \textbf{Safe-Driving Zone:} The period of safe, nominal driving. This zone spans from the beginning of the sequence to a predefined critical horizon frame, denoted as $t \in [0, t_h)$.
    \item \textbf{Critical Zone:} The period immediately preceding the collision, during which the system is expected to anticipate the event. This zone is defined as $t \in [t_h, t_{\text{acc}})$.
\end{itemize}
For a safe video containing no accidents, the horizon and accident frames are equivalent to the total video length, $t_h = t_{\text{acc}} = T$, \ie, the critical zone is empty. 
Ideally, an accident anticipation system must trigger warnings within the critical zone ($t \ge t_h$) to enable timely reactions:
\begin{equation}
a_t =
\begin{cases}
    1 & t \ge t_h \\
    0 & \text{otherwise}
\end{cases}
\end{equation}
Equally, the system should strictly suppress warnings during the safe-driving zone ($t < t_h$) to prevent excessive false alarms. We propose a novel evaluation metric in \secref{sec:metric} that considers both anticipation capacity in the critical zone and a low false alarm rate in the safe-driving zone.

\subsection{Accident anticipation}
\label{subsec:architecture}
To capture the dynamics leading to an accident, we construct the state $\bs_t \in \mathbb{R}^{N \times H \times W \times 3}$ as a window of $N$ recent frames. We then encode $\bs_t$ into a set of spatio-temporal tokens $\cZ_t$ using a video foundation model (FoMo):
\begin{equation}
    \cZ_t = \Phi_{\text{FoMo}}(\bs_t), \quad \cZ_t \in \mathbb{R}^{K \times D}.
\end{equation}
where $K$ denotes the number of tokens and $D$ is the hidden dimension. To leverage the pre-training of the FoMo on large-scale datasets, we keep most layers frozen and finetune only the final layers.

To obtain a global representation of the segment, we apply average pooling over the $K$ tokens, resulting in $\bz_t$. We then process $\bz_t$ with a lightweight MLP head to predict an accident score $\hat{a}_t$:
\begin{eqnarray}
    \bz_t &=& \Phi_{\text{pool}}(\cZ_t), \quad \bz_t \in \mathbb{R}^{D} \\
    \hat{a}_t &=& \Phi_{\text{MLP}} (\bz_t).
    \label{eq:mlp}
\end{eqnarray}
Segments in the critical zone are treated as positive examples, while negative samples are drawn from the safe-driving zone. See the model and the head architecture in \figref{fig:model_overview}.

\subsection{Continuous risk estimation}
\label{subsec:progress_estimation}

Similar to prior work in accident anticipation~\cite{zhao2025accident}, the model described in the previous section can be trained using a BCE loss. However, this formulation does not provide an explicit ordering between segments in terms of their proximity to an accident. To address this, we reformulate accident anticipation as a monotonically increasing function over time.

We define a temporal risk level $r_t$, which serves as our progress estimation target:
\begin{equation}
    \label{eq:risk_target_exp}
    \tau = \frac{t - t_h}{t_{\text{acc}} - t_h}, \quad\quad
    r_t =
    \begin{cases}
        0, & \text{if } t < t_h, \\[1mm]
        \dfrac{1 - \exp(-\alpha \tau)}
        {1 - \exp(-\alpha)},
        & \text{if } t_h \leq t < t_{\text{acc}}.
    \end{cases}
\end{equation}
%

% Requires: \usepackage{pgfplots}
%          \pgfplotsset{compat=1.18}
% Put risk_targets_better.csv in the same folder as this .tex file.

\begin{wrapfigure}{r}{0.4\textwidth}
\centering
% \vspace{-15pt}

\begin{tikzpicture}
\begin{axis}[
    width=1\linewidth,
    height=0.9\linewidth,
    xlabel={Time-to-collision (Frame $t \times 10$ FPS)},
    ylabel={Risk level ($r_t$)},
    xmin=0, xmax=20,
    x dir=reverse,
    ymin=0, ymax=1.02,
    xtick={20,15,10,5,0},
    ytick={0,0.2,0.4,0.6,0.8,1.0},
    grid=both,
    grid style={line width=.1pt, draw=gray!20},
    major grid style={line width=.2pt, draw=gray!35},
    legend style={
        at={(0.5,-0.28)},
        anchor=north,
        legend columns=5,
        font=\scriptsize,
        draw=none,
        /tikz/every even column/.append style={column sep=0.25cm}
    },
    tick label style={font=\small},
    label style={font=\small},
]

% Linear baseline
\addplot[black, very thick, dashed, mark=none]
table[x=ttc, y=linear, col sep=comma]
{./figures/appendix/linear_exp_under_above_alphas.csv};
% \addlegendentry{Linear}

% Under-linear exponentials: delayed risk growth
\addplot[blue!35!cyan, thick, mark=none]
table[x=ttc, y=exp_under_alpha_1, col sep=comma]
{./figures/appendix/linear_exp_under_above_alphas.csv};
% \addlegendentry{$\alpha=-1$}

\addplot[blue!55!cyan, thick, mark=none]
table[x=ttc, y=exp_under_alpha_2, col sep=comma]
{./figures/appendix/linear_exp_under_above_alphas.csv};
% \addlegendentry{$\alpha=-2$}

\addplot[blue!75!cyan, thick, mark=none]
table[x=ttc, y=exp_under_alpha_3, col sep=comma]
{./figures/appendix/linear_exp_under_above_alphas.csv};
% \addlegendentry{$\alpha=-3$}

\addplot[blue!80!black, thick, mark=none]
table[x=ttc, y=exp_under_alpha_4, col sep=comma]
{./figures/appendix/linear_exp_under_above_alphas.csv};
%\addlegendentry{$\alpha=-4$}

\addplot[blue!65!black, thick, mark=none]
table[x=ttc, y=exp_under_alpha_5, col sep=comma]
{./figures/appendix/linear_exp_under_above_alphas.csv};
%\addlegendentry{$\alpha=-5$}

\addplot[blue!40!black, thick, mark=none]
table[x=ttc, y=exp_under_alpha_10, col sep=comma]
{./figures/appendix/linear_exp_under_above_alphas.csv};
%\addlegendentry{$\alpha=-10$}

% Above-linear exponentials: early risk growth
\addplot[orange!70!red, thick, mark=none]
table[x=ttc, y=exp_above_alpha_1, col sep=comma]
{./figures/appendix/linear_exp_under_above_alphas.csv};
%\addlegendentry{$\alpha=1$}

\addplot[orange!85!red, thick, mark=none]
table[x=ttc, y=exp_above_alpha_2, col sep=comma]
{./figures/appendix/linear_exp_under_above_alphas.csv};
%\addlegendentry{$\alpha=2$}

\addplot[red!70!black, very thick, mark=none]
table[x=ttc, y=exp_above_alpha_3, col sep=comma]
{./figures/appendix/linear_exp_under_above_alphas.csv};
%\addlegendentry{$\alpha=3$}

\addplot[red!80!black, thick, mark=none]
table[x=ttc, y=exp_above_alpha_4, col sep=comma]
{./figures/appendix/linear_exp_under_above_alphas.csv};
%\addlegendentry{$\alpha=4$}

\addplot[red!90!black, thick, mark=none]
table[x=ttc, y=exp_above_alpha_5, col sep=comma]
{./figures/appendix/linear_exp_under_above_alphas.csv};
%\addlegendentry{$\alpha=5$}

\addplot[magenta!70!black, thick, mark=none]
table[x=ttc, y=exp_above_alpha_10, col sep=comma]
{./figures/appendix/linear_exp_under_above_alphas.csv};
% \addlegendentry{$\alpha=10$}

\end{axis}
\end{tikzpicture}
\caption{\textbf{Varying $\alpha$.} Continuous risk curves with fast-increasing (red side) and slow-increasing (blue side) behavior obtained by varying $\alpha$. The linear case is shown with a dashed line.}
\label{fig:risk-targets-wrap}
\vspace{-7mm}
\end{wrapfigure}
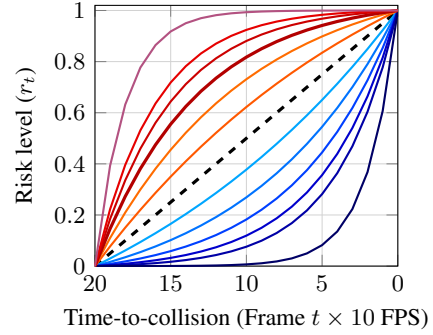

The risk $r_t$ is strictly $0$ in the safe-driving zone. After the critical horizon $t_h$, it increases within the critical zone as a function of the normalized time $\tau$, approaching $1$ at $t_{\text{acc}}$. We explore both fast- and slow-increasing variants with different growth behavior by adjusting the non-zero curvature parameter, $\alpha$, as shown in \figref{fig:risk-targets-wrap}.

We introduce an additional MLP head on top of the FoMo features, as in \eqnref{eq:mlp}, to predict the risk progress $\hat{r}_t$. This head is trained to approximate $r_t$ using a smooth $\mathcal{L}_{1}$ loss.

\boldparagraph{Preference estimation}
To further encourage temporal ordering between segments, we introduce an auxiliary preference ranking objective. Given a sampled segment, we randomly select another segment from the critical zone of the same video with a higher risk level, \ie, closer to the accident. Using another MLP head, we predict risk scores for both segments and enforce the higher-risk segment to receive a higher score via a margin ranking loss.

We jointly train the model to anticipate accidents, estimate the continuous risk progression toward them, and rank the temporal closeness of states to the accident. The overall objective $\mathcal{L}$ is defined as a weighted combination of binary classification $(\mathcal{L}_{\text{BCE}})$, progress estimation $(\mathcal{L}_1)$, and preference ranking $(\mathcal{L}_{\text{rank}})$ losses:
\begin{equation}
\mathcal{L} = w_{\text{acc}}~\mathcal{L}_{\text{BCE}} +
w_{\text{prog}}~\mathcal{L}_1 +
w_{\text{pref}}~\mathcal{L}_{\text{rank}}
\label{eq:overall_obj}
\end{equation}
\section{Beyond AUC and TTA: Separation score  for global risk evaluation}
\label{sec:metric}

\subsection{Current state of evaluation}
\label{sec:current_metrics} 

\textbf{AUC} is a standard metric for accident anticipation; however, it requires defining positive and negative samples. Positives are defined at 1.5, 1.0, 0.5, and 0.0 seconds before the accident, where the last corresponds to the moment of impact. Negatives are sampled at a certain temporal distance from the accident, but not too far, \ie, from frames up to 0.5~s before the anomaly onset.

\textbf{(mean) Time-To-Accident (mTTA)} evaluates how early a model anticipates an accident. It is computed based on the first significant prediction peak after the ground-truth anomaly onset ($t_{ai}$), indicating the earliest time at which the model detects the impending crash.

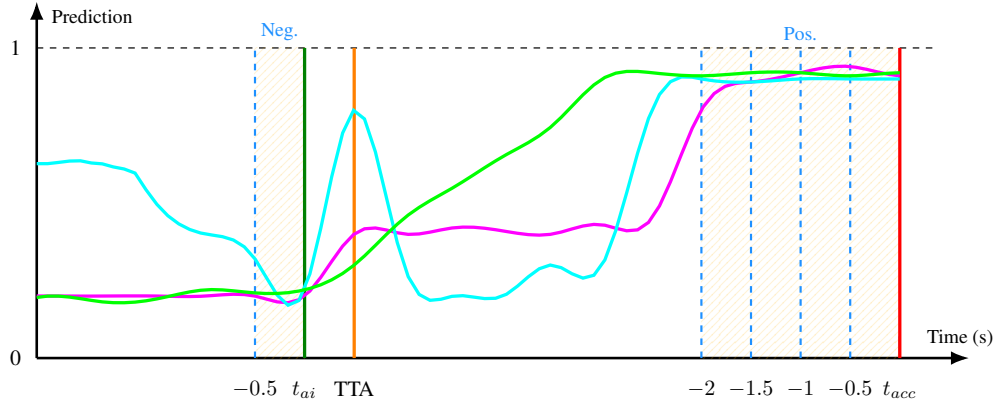
\begin{figure}[t]
\centering

\resizebox{0.95\columnwidth}{!}{
\begin{tikzpicture}

% Axes
\draw[-{Latex},line width=1.2pt] (0.48,0) -- (15.52,0) node[above left,font=\small,align=left,yshift=-7pt,xshift=15pt] {Time (s)\\};
\draw[-{Latex},line width=1.2pt] (0.48,0) -- (0.48,5.75) node[below right,font=\small,xshift=3pt,yshift=-0.1pt] {Prediction};

% y-axis labels
\node[left,font=\normalsize] at (0.35,5.0) {1};
\node[left,font=\normalsize] at (0.35,0.0) {0};

% Reference lines
\draw[black,dashed,line width=0.8pt,opacity=0.7] (0.48,5.0) -- (15.,5.0);

% Tick marks (only at t_ai and t_co)
\draw[black,line width=1pt] (4.800,0) -- (4.800,0.35);
\draw[black,line width=1pt] (14.400,0) -- (14.400,0.35);

% Hatched scanning regions
\fill[pattern=north east lines,pattern color=orange!60!yellow,opacity=0.4] (4.000,0) rectangle (4.800,5.0);
\fill[pattern=north east lines,pattern color=orange!60!yellow,opacity=0.4] (11.200,0) rectangle (14.400,5.0);

% Blue dashed lines and labels
\draw[dodgerblue,dashed,line width=1pt] (13.600,0) -- (13.600,5.0);
\node[below,font=\normalsize] at (13.600,-0.25) {$-0.5$};
\draw[dodgerblue,dashed,line width=1pt] (12.800,0) -- (12.800,5.0);
\node[below,font=\normalsize] at (12.800,-0.25) {$-1$};
\draw[dodgerblue,dashed,line width=1pt] (12.000,0) -- (12.000,5.0);
\node[below,font=\normalsize] at (12.000,-0.25) {$-1.5$};
\draw[dodgerblue,dashed,line width=1pt] (11.200,0) -- (11.200,5.0);
\node[below,font=\normalsize] at (11.200,-0.25) {$-2$};

% Neg line
\draw[dodgerblue,dashed,line width=1pt] (4.000,0) -- (4.000,5.0);
\node[dodgerblue,above,font=\small] at (4.4,5.0) {Neg.};
\node[dodgerblue,above,font=\small] at (12.8,5.0) {Pos.};
\node[below,font=\normalsize] at (4.000,-0.25) {$-0.5$}; %{$t_{ai}-0.5$};

% Vertical lines at t_ai (green) and t_co (red)
\draw[green!50!black,line width=1.5pt] (4.800,0) -- (4.800,5.0);
\draw[red,line width=1.5pt] (14.400,0) -- (14.400,5.0);

\draw[orange,line width=1.5pt] (5.600,0) -- (5.600,5.0);

% x-axis labels
\node[below,font=\normalsize] at (4.800,-0.25) {$t_{ai}$};
\node[below,font=\normalsize] at (14.400,-0.25) {$t_{acc}$};

\node[below,font=\normalsize] at (5.600,-0.25) {TTA};

% Curves
\draw[mplmagenta,line width=1.5pt,line cap=butt] plot coordinates {(0.480,1.001) (0.656,1.000) (0.832,1.000) (1.009,1.000) (1.185,0.999) (1.361,0.999) (1.537,1.000) (1.713,1.001) (1.890,1.002) (2.066,1.002) (2.242,1.002) (2.418,1.000) (2.594,0.996) (2.771,0.993) (2.947,0.992) (3.123,0.996) (3.299,1.007) (3.475,1.020) (3.652,1.029) (3.828,1.024) (4.004,0.999) (4.180,0.952) (4.356,0.906) (4.533,0.893) (4.709,0.941) (4.885,1.080) (5.061,1.301) (5.237,1.558) (5.414,1.805) (5.590,1.992) (5.766,2.084) (5.942,2.097) (6.118,2.065) (6.295,2.020) (6.471,1.994) (6.647,2.001) (6.823,2.030) (6.999,2.066) (7.176,2.097) (7.352,2.111) (7.528,2.109) (7.704,2.093) (7.881,2.069) (8.057,2.040) (8.233,2.012) (8.409,1.990) (8.585,1.982) (8.762,1.994) (8.938,2.031) (9.114,2.082) (9.290,2.129) (9.466,2.155) (9.643,2.142) (9.819,2.093) (9.995,2.052) (10.171,2.067) (10.347,2.187) (10.524,2.453) (10.700,2.840) (10.876,3.280) (11.052,3.705) (11.228,4.047) (11.405,4.264) (11.581,4.380) (11.757,4.429) (11.933,4.446) (12.109,4.461) (12.286,4.487) (12.462,4.522) (12.638,4.562) (12.814,4.603) (12.990,4.644) (13.167,4.678) (13.343,4.700) (13.519,4.705) (13.695,4.688) (13.871,4.653) (14.048,4.611) (14.224,4.573) (14.400,4.550)};
\draw[mplcyan,line width=1.5pt,line cap=butt] plot coordinates {(0.480,3.134) (0.656,3.134) (0.832,3.154) (1.009,3.176) (1.185,3.180) (1.361,3.145) (1.537,3.131) (1.713,3.082) (1.890,3.048) (2.066,2.982) (2.242,2.714) (2.418,2.478) (2.594,2.295) (2.771,2.163) (2.947,2.073) (3.123,2.016) (3.299,1.985) (3.475,1.957) (3.652,1.903) (3.828,1.793) (4.004,1.594) (4.180,1.304) (4.356,1.017) (4.533,0.852) (4.709,0.928) (4.885,1.356) (5.061,2.091) (5.237,2.922) (5.414,3.629) (5.590,3.994) (5.766,3.855) (5.942,3.324) (6.118,2.595) (6.295,1.861) (6.471,1.313) (6.647,1.026) (6.823,0.931) (6.999,0.948) (7.176,0.995) (7.352,1.006) (7.528,0.983) (7.704,0.959) (7.881,0.964) (8.057,1.029) (8.233,1.159) (8.409,1.311) (8.585,1.440) (8.762,1.500) (8.938,1.458) (9.114,1.359) (9.290,1.289) (9.466,1.335) (9.643,1.582) (9.819,2.056) (9.995,2.665) (10.171,3.303) (10.347,3.863) (10.524,4.249) (10.700,4.455) (10.876,4.532) (11.052,4.528) (11.228,4.495) (11.405,4.469) (11.581,4.453) (11.757,4.447) (11.933,4.448) (12.109,4.455) (12.286,4.466) (12.462,4.479) (12.638,4.492) (12.814,4.501) (12.990,4.505) (13.167,4.505) (13.343,4.504) (13.519,4.501) (13.695,4.499) (13.871,4.499) (14.048,4.499) (14.224,4.499) (14.400,4.500)};
\draw[mpllime,line width=1.5pt,line cap=butt] plot coordinates {(0.480,0.977) (0.656,0.999) (0.832,0.999) (1.009,0.983) (1.185,0.958) (1.361,0.930) (1.537,0.906) (1.713,0.893) (1.890,0.892) (2.066,0.902) (2.242,0.923) (2.418,0.954) (2.594,0.991) (2.771,1.031) (2.947,1.067) (3.123,1.093) (3.299,1.104) (3.475,1.099) (3.652,1.084) (3.828,1.066) (4.004,1.050) (4.180,1.041) (4.356,1.042) (4.533,1.054) (4.709,1.081) (4.885,1.122) (5.061,1.181) (5.237,1.261) (5.414,1.364) (5.590,1.492) (5.766,1.646) (5.942,1.817) (6.118,1.992) (6.295,2.159) (6.471,2.306) (6.647,2.429) (6.823,2.537) (6.999,2.636) (7.176,2.736) (7.352,2.843) (7.528,2.955) (7.704,3.069) (7.881,3.179) (8.057,3.282) (8.233,3.377) (8.409,3.474) (8.585,3.584) (8.762,3.717) (8.938,3.880) (9.114,4.062) (9.290,4.244) (9.466,4.404) (9.643,4.524) (9.819,4.593) (9.995,4.621) (10.171,4.621) (10.347,4.606) (10.524,4.587) (10.700,4.571) (10.876,4.558) (11.052,4.551) (11.228,4.550) (11.405,4.557) (11.581,4.569) (11.757,4.582) (11.933,4.596) (12.109,4.606) (12.286,4.611) (12.462,4.612) (12.638,4.608) (12.814,4.599) (12.990,4.585) (13.167,4.570) (13.343,4.557) (13.519,4.550) (13.695,4.552) (13.871,4.563) (14.048,4.578) (14.224,4.592) (14.400,4.600)};

\end{tikzpicture}}

% \caption{}
% \label{fig:bottom}
% \end{subfigure}

\caption{\textbf{Problem with current evaluation metrics.}
%$t_{ai}$ denotes the anomaly onset time, and $t_{acc}$ denotes the collision time. 
%Standard AUC evaluation considers only local windows: positives are sampled near the accident, \ie, $t_{acc}-2$, while negatives are sampled near anomaly onset, \ie, $t_{ai}-0.5$, shown by the yellow hatched regions. 
%The {\textcolor{mplcyan}{{cyan}}}, {\textcolor{mplmagenta}{{magenta}}}, and {\textcolor{mpllime}{{green}}} curves obtain nearly identical AUC values, close to the optimal value of 1. Similarly, all three achieve the same TTA, as they increase relative to the pre-anomaly period at approximately the same time. However, their global risk behavior differs substantially. The {\textcolor{mplcyan}{{cyan}}} curve produces false alarms before the anomaly, while the {\textcolor{mplmagenta}{{magenta}}} curve remains mostly unchanged as the crash approaches. In contrast, the {\textcolor{mpllime}{{green}}} curve shows the desired behavior: low risk before $t_{ai}$, increasing risk after anomaly onset, and progressively higher risk as $t_{acc}$ approaches.}
Standard AUC evaluates predictions only within local temporal windows: negatives are sampled around anomaly onset ($t_{ai}-0.5$) and positives close to the accident ($t_{acc}-2$), illustrated by the yellow hatched regions. As a result, the {\textcolor{mplcyan}{cyan}}, {\textcolor{mplmagenta}{magenta}}, and {\textcolor{mpllime}{green}} curves achieve similarly high AUC (near 1) and identical TTA, since they rise relative to the pre-anomaly region at comparable times ({\textcolor{orange}{orange}} line). However, their global behavior differs significantly. The {\textcolor{mplcyan}{cyan}} curve produces early false alarms before $t_{ai}$, while the {\textcolor{mplmagenta}{magenta}} curve shows weak progression toward the accident. The {\textcolor{mpllime}{green}} curve exhibits the desired behavior: low risk before $t_{ai}$, a clear increase after anomaly onset, and steadily rising risk as $t_{acc}$ ({\textcolor{red}{red}} line) approaches.}
\label{fig:metrics}
\end{figure}

\boldquestion{What is missing in current metrics?}
Although these metrics provide useful measures of anticipation performance, they evaluate predictions only at selected temporal intervals. As shown in \figref{fig:metrics}, this fixed-interval evaluation focuses on a narrow temporal window when comparing different methods. Moreover, these metrics often rely on dataset-specific thresholds that must be tuned for each dataset.
 
This local nature of evaluation can obscure important differences in model behavior over time. In particular, these metrics do not capture the overall shape of the predicted risk curve. As illustrated in \figref{fig:metrics}, three different prediction curves can yield very similar AUC and TTA values because evaluation is restricted to local temporal windows. However, in accident anticipation, evaluating only a few isolated time snippets is insufficient. From a safety perspective, a reliable model should avoid producing high-risk predictions before the anomaly occurs, thereby minimizing false alarms, while increasing its predictions only after visual cues of an upcoming accident become observable.

\subsection{A new separation metric for accident anticipation}

An ideal model should assign low scores to pre-anomaly regions, \ie, up to the anomaly onset $t_{\text{ai}}$, and high scores afterward, \ie, from $t_{\text{ai}}$ to $t_{\text{acc}}$, creating a clear separation between normal and anomalous temporal segments. Motivated by this observation, we propose a new metric, the \textit{Separation Score}, for the global evaluation of risk prediction curves. The proposed metric compares the area under the predicted risk curve before and after the anomaly onset and combines them into a single score:
\begin{equation}
s_{\text{pre}} = \frac{1}{N_{\mathrm{pre}}}\sum_{i=1}^{t_{\mathrm{ai}} -1} \hat{a}_i, \quad \quad
s_{\text{post}} = \frac{1}{N_{\mathrm{post}}}\sum_{j=t_{\mathrm{ai}}}^{t_{\mathrm{acc}}} \hat{a}_j, \quad \quad
s = \frac{1 - s_{\text{pre}} + s_{\text{post}}}{2}.
\end{equation}
where $\hat{a}_i$ denotes the predicted risk score at time step $i$, and $N_{\mathrm{pre}}$ and $N_{\mathrm{post}}$ denote the number of pre- and post-anomaly predictions, respectively. We compute the average risk before and after the anomaly onset as $s_{\text{pre}}$ and $s_{\text{post}}$, and combine them into a normalized separation score $s \in [0,1]$, where higher values indicate better performance. The ideal case $s=1$ corresponds to zero risk before the anomaly and unit risk after its onset, indicating perfect separation.

\section{Experiments}
\label{sec:exp}

\subsection{Experimental setup}
\label{sec:setup}

\boldparagraph{Datasets} We conduct experiments on two recent crash anticipation benchmarks: MM-AU~\cite{fang2024abductive} and Nexar~\cite{moura2025nexar}. 
MM-AU contains two splits: DADA~\cite{fang2021dada}, a driver-attention-oriented dashcam accident dataset with 1,771 training and 198 test videos recorded at 30 FPS, and CAP~\cite{fang2022cognitive}, a larger-scale accident understanding dataset with richer scenario diversity, containing 8,959 training and 800 test videos with varying FPS values. 
We train and evaluate our models separately on the DADA and CAP splits. 
The Nexar Dashcam Collision Prediction Dataset contains 3,000 training videos and 1,354 test videos, with balanced accident and non-accident samples. 
Since the challenge permits the use of public datasets, we first train our model on CAP and then fine-tune it on Nexar.

\boldparagraph{Implementation details}
We preprocess input videos by resizing frames to $224 \times 224$ and subsampling them at 10 FPS. Based on training data statistics, we set the horizon parameter $t_h$ to 2 seconds. During training, we sample 5-frame clips from segments preceding the accident frame $t_{\text{acc}}$. To avoid an overwhelming number of negative samples, we limit the number of negative clips per video to 50. 
We set $\alpha = 3$ in the risk level function in \eqnref{eq:risk_target_exp}.
For the preference loss, we sample more risky windows from a temporal range of 0.5 to 1.5 seconds after the current segment, moving toward the accident. During inference, we apply the model in a sliding-window fashion over the video. 
Our default video encoder is the large version of VideoMAE~\cite{tong2022videomae}, pre-trained on Kinetics-400~\cite{kay2017kinetics}. The task-specific heads consists of two linear layers with GELU activations~\cite{hendrycks2016gelu}. We train the task heads together with the last four layers of the video encoder for one epoch using a batch size of 32 on a single NVIDIA A100 GPU. The learning rates are set to $1\mathrm{e}{-5}$ for the video encoder and $1\mathrm{e}{-4}$ for the task heads. We set loss weights in \eqnref{eq:overall_obj} as follows: $w_{\text{acc}} = 1.0$, $w_{\text{prog}} = 10.0$, and $w_{\text{pref}} = 0.1$.

\boldparagraph{Metrics}
We evaluate our models using the metrics described in \cref{sec:current_metrics}. We report the \emph{Separation Score} only for our method, as pretrained models from prior work are not publicly available. Additionally, we report results under an FPR $\leq 0.1$ constraint, following \cite{zhao2025accident}, since excessive false alarms can reduce trust in the system and limit its practical usefulness.

\subsection{Quantitative results}

%%%% Table CAP Results.
\begin{table}[t]
\caption{\textbf{Results on MM-AU.} We report results on the CAP~\cite{fang2022cognitive} (top) and DADA~\cite{fang2021dada} (bottom) splits of MM-AU~\cite{fang2024abductive}. Our method consistently outperforms prior work across almost all metrics by substantial margins, with improvements becoming more pronounced closer to the accident.}
\label{tab:cap_dada_sota}

\begin{center}
%\resizebox{1.0\columnwidth}{!}{
\begin{tabular}{llcccccc}
\toprule % <-- Toprule here
& {Method} & AUC$_{0.0s}^{0.1}$ & AUC$_{0.5s}^{0.1}$ & AUC$_{1.0s}^{0.1}$ &  AUC$_{1.5s}^{0.1}$ & mAUC$^{0.1}$ & mTTA$^{0.1}$ \\
\midrule % <-- Midrule here

\multirow{6}{*}{\rotatebox{90}{CAP}} & 
CAP~\cite{fang2022cognitive}    & 0.042 &  0.040 & 0.030 & 0.037 & 0.036 & 0.637 \\
& DRIVE~\cite{bao2021drive}       &  0.129 & 0.117 & 0.108 & 0.123 & 0.116 & 0.395 \\
& DSTA~\cite{karim2022dynamic}    & 0.559 &  0.386 & 0.282 & 0.191 & 0.286 & 0.804 \\
& GSC~\cite{wang2023gsc}          &   0.609 & 0.418 & 0.297 & 0.199 & 0.305 & 0.817 \\
& TOP~\cite{zhao2025accident}     &  0.838 & 0.675 & 0.398 & \textbf{0.214}& 0.429 & 0.864 \\

% \midrule % <-- Midrule here
\cmidrule(lr){2-8} % <-- Midrule here

& {\ours} & \textbf{0.887}&\textbf{0.775} & \textbf{0.465} & 0.201 & \textbf{0.481}& \textbf{1.125}\\

\midrule % <-- Midrule here
\midrule % <-- Midrule here

\multirow{6}{*}{\rotatebox{90}{DADA}} & 
CAP~\cite{fang2022cognitive}    & 0.032 &   0.037 & 0.067 & 0.064 & 0.056 & 0.496 \\
& DRIVE~\cite{bao2021drive}       & 0.101 &  0.063 & 0.077 & 0.088 & 0.076 & 0.226 \\
& DSTA~\cite{karim2022dynamic}    &   0.473 & 0.328 & 0.221 & 0.135 & 0.228 & 0.695 \\
& GSC~\cite{wang2023gsc}          &  0.514 & 0.350 & 0.238 & 0.139 & 0.242 & 0.703 \\
& TOP~\cite{zhao2025accident}     & 0.790 &  0.567 & 0.288 & 0.140 & 0.332 & 0.885 \\

\cmidrule(lr){2-8} % <-- Midrule here % <-- Midrule here

& {\ours} & \textbf{0.861} & \textbf{0.751} & \textbf{0.398} & \textbf{0.170}&\textbf{0.440}& \textbf{1.069} \\

\bottomrule % <-- Bottomrule here
\end{tabular}%}

% \begin{tablenotes} %[para]
%     \item 
%     \footnotesize Baseline results are directly taken from TOP~\cite{zhao2025accident}.
% \end{tablenotes}

%\vspace{-5mm}
\end{center}
\end{table}

\boldparagraph{MM-AU \cite{fang2024abductive}}
We compare our method with prior work in \tabref{tab:cap_dada_sota} on the CAP~\cite{fang2022cognitive} and DADA~\cite{fang2021dada} splits of MM-AU, using the baseline results reported in \cite{zhao2025accident}. Our method surpasses the current state-of-the-art TOP~\cite{zhao2025accident} by over $\mathbf{10\%}$ on CAP and nearly $\mathbf{30\%}$ on DADA in terms of mAUC. 
The improvement on CAP further increases to almost $\mathbf{15\%}$ at shorter horizons (AUC$_{0.5s}$), where positive frames are closer to the collision, highlighting the effectiveness of our risk formulation in emphasizing near-accident scenarios. 
Our method also anticipates accidents earlier than prior approaches, as reflected by consistent gains in mTTA, improving by \textbf{0.25~s} on CAP and \textbf{0.18~s} on DADA.

\begin{wraptable}{r}{0.3\textwidth}
\vspace{-5mm}
\caption{\textbf{Results on Nexar.}}
\vspace{-3.5mm}
\begin{center}
% \resizebox{0.33\columnwidth}{!}{
\begin{tabular}{lc}
\toprule % <-- Toprule here
Method & mAP \\
\midrule % <-- Midrule here
Runner-up    & 0.875  \\
Winner    & 0.885  \\
% \midrule
\ours       & \textbf{0.895} \\

\bottomrule % <-- Bottomrule here
\end{tabular}%}
\end{center}
\label{tab:nexar_map}
\vspace{-5mm}
\end{wraptable}

\boldparagraph{Nexar \cite{moura2025nexar}}
We report our results on Nexar in \tabref{tab:nexar_map}, comparing against the winner and runner-up as listed on the challenge leaderboard\footnote{https://www.kaggle.com/competitions/nexar-collision-prediction/leaderboard}.
Our method demonstrates strong performance on Nexar, outperforming the previous winner by $\mathbf{+1}$ mAP. 
While Nexar officially uses mAP as the primary evaluation metric, we additionally report results with our standard metrics in \appref{app:nexar_results}.

%%%% Table CAP Results.
\begin{table}[h]
\caption{\textbf{Cross-dataset generalization.} We train our model on the largest subset, CAP, and evaluate it on DADA and Nexar. As indicated by the performance differences (diff.; {\textcolor{green!30!black}{increase}} and {\textcolor{MyRed}{decrease}}) relative to models trained on each dataset’s own training set, our model generalizes well to DADA, with a slight performance drop on Nexar due to a larger domain gap.}
\label{tab:zero_shot_eval}

\begin{center}
%\resizebox{1.0\columnwidth}{!}{
\begin{tabular}{llrrrrrrr}
\toprule % <-- Toprule here
{Train} & Test & AUC$_{0.0s}^{0.1}$ & AUC$_{0.5s}^{0.1}$ & AUC$_{1.0s}^{0.1}$ &  AUC$_{1.5s}^{0.1}$ & mAUC$^{0.1}$ & mTTA$^{0.1}$  & $s$\\
\midrule % <-- Midrule here

\multirow{4}{*}{CAP} & DADA & 0.897 & 0.799 & 0.413 & 0.178 & 0.464 &  1.155 & 0.760 \\
\multicolumn{2}{r}{\scriptsize diff. }
& {\scriptsize\textcolor{green!30!black}{0.036}}
& {\scriptsize\textcolor{green!30!black}{0.048}}
& {\scriptsize\textcolor{green!30!black}{0.015}}
& {\scriptsize\textcolor{green!30!black}{0.008}}
& {\scriptsize\textcolor{green!30!black}{0.024}}
& {\scriptsize\textcolor{green!30!black}{0.086}}  
& {\scriptsize\textcolor{green!30!black}{0.020}}\\

 & Nexar & - & 0.529 & 0.336 & 0.218 & 0.361 & 1.020 & 0.650 \\
 \multicolumn{2}{r}{\scriptsize diff. }
& {\scriptsize\textcolor{black}{-}}
& {\scriptsize\textcolor{MyRed}{0.082}}
& {\scriptsize\textcolor{MyRed}{0.236}}
& {\scriptsize\textcolor{MyRed}{0.151}}
& {\scriptsize\textcolor{MyRed}{0.149}}
& {\scriptsize\textcolor{MyRed}{0.265}}
& {\scriptsize\textcolor{MyRed}{0.049}} \\

\bottomrule % <-- Bottomrule here
\end{tabular}%}
\end{center}
\end{table}

\boldparagraph{Cross-dataset generalization}
The CAP~\cite{fang2022cognitive} subset contains substantially more training samples than DADA~\cite{fang2021dada} and Nexar~\cite{moura2025nexar}, providing a richer source of supervision. To evaluate cross-dataset generalization, we train our model on CAP and directly test it on DADA and Nexar without any further fine-tuning. Results are reported in \tabref{tab:zero_shot_eval}, together with performance differences relative to models trained and evaluated on each target dataset. 

Training on CAP transfers well to DADA, yielding improvements across most metrics. This indicates that the scale and diversity of CAP enable the model to learn transferable accident-related cues. In contrast, performance degrades slightly on Nexar, likely due to a larger domain gap in visual appearance and driving conditions, \eg, differences in geography, camera setup, and traffic patterns.

The link in the abstract provides \emph{qualitative results} and \emph{failure cases}.
Failures mainly occur due to poor visibility, such as snow or strong frontal sunlight, degrading perception quality.

%%%% Table CAP Loss Ablation Results.
\begin{table}[t]
\caption{\textbf{Ablation on progress and preference estimation.} Compared to the BCE baseline, progress estimation (Prog.) significantly improves performance across all metrics. Adding preference estimation (Pref.) provides slight additional gains, achieving the best overall results.}
\label{tab:abl_losses}

\begin{center}
\resizebox{1.0\columnwidth}{!}{
\begin{tabular}{lll|ccccccc}
\toprule % <-- Toprule here
BCE & Prog. & Pref. & AUC$_{0.0s}^{0.1}$ & AUC$_{0.5s}^{0.1}$ & AUC$_{1.0s}^{0.1}$ &  AUC$_{1.5s}^{0.1}$ & mAUC$^{0.1}$ & mTTA$^{0.1}$  & $s$ \\
\midrule % <-- Midrule here
\cmark &  \xmark & \xmark &  0.807	& 0.690	& 0.392	& 0.176	& 0.419 & 1.105 & 0.694 \\
\cmark  & \cmark &  \xmark & 0.884	& 0.773	& 0.463	& 0.197	& 0.477 & 1.128 & 0.723 \\
\cmark  & \cmark &  \cmark & 0.887	& 0.775	& 0.465	& 0.201	& 0.481 & 1.125 & 0.724 \\

\bottomrule % <-- Bottomrule here
\end{tabular}}
\end{center}
\end{table}

\subsection{Ablation study}
\label{sec:ablations}

We perform ablations on the CAP subset using a vanilla model trained with BCE loss as the baseline. Our analysis studies the contributions of progress and preference estimation, the effect of different risk functions, and the ability of the proposed Separation Score to capture differences in the global behavior of models. 
We further evaluate the robustness of our model under varying FPR in \appref{app:varying_fpr} and the impact of different backbone choices in \appref{app:varying_backbone}.

%%%% Table CAP Results.
\begin{table}[b]
\centering
% \small
\vspace{-2mm}
\caption{\textbf{Results with varying risk functions.} Fast-increasing risk functions with positive $\alpha$ improve longer-horizon anticipation, \ie, $\ge 1$ sec, while slow-increasing functions with negative $\alpha$ perform better at shorter horizons, \ie, $< 1$ sec. $\alpha=3$ achieves the best overall performance.}
\label{tab:risk_func_abl}

\begin{center}
%\resizebox{1.0\columnwidth}{!}{
\begin{tabular}{ccccccccc}
\toprule % <-- Toprule here
{Increase} & $\alpha$ & AUC$_{0.0s}^{0.1}$ & AUC$_{0.5s}^{0.1}$ & AUC$_{1.0s}^{0.1}$ &  AUC$_{1.5s}^{0.1}$ & mAUC$^{0.1}$ & mTTA$^{0.1}$ & $s$ \\
\midrule % <-- Midrule here

\multirow{6}{*}{Fast} &  \textcolor{magenta!70!black}{$10$}     &  0.858	& 0.750	 & 0.459	& 0.203  & 0.471  & 1.120  & 0.722  \\
&  \textcolor{red!90!black}{$5$}      &  0.877	& 0.768	 & \textbf{0.467}	& \textbf{0.204}	 & 0.480  & \textbf{1.125}  & 0.724  \\
&  \textcolor{red!80!black}{$4$}      &  0.882	& 0.772	 & 0.466	& \textbf{0.204}	 & \textbf{0.481}  & \textbf{1.125}  & 0.724  \\
&  \textcolor{red!70!black}{$3$}      &  0.887	& 0.775	 & 0.465 & 0.201	 & \textbf{0.481}  & \textbf{1.125}  & 0.724  \\
&  \textcolor{orange!85!red}{$2$}     &  0.893	& \textbf{0.778}	 & 0.461	& 0.198	 & 0.479  & 1.124  & 0.725  \\
&  \textcolor{orange!70!red}{$1$}     &  0.898	& \textbf{0.778}	 & 0.454	& 0.193	 & 0.475  & {1.122}  & 0.726  \\
\midrule
Linear   & \textcolor{black}{N/A }            &  0.901	& \textbf{0.778}	 & 0.442	& 0.190	 & 0.470  & 1.118  & 0.726  \\

\midrule
\multirow{6}{*}{Slow} & 
 \textcolor{blue!35!cyan}{$-1$}      &  \textbf{0.902}	& 0.776	 & 0.435	& 0.187	 & 0.466  & 1.115  & \textbf{0.727}  \\
& \textcolor{blue!55!cyan}{$-2$}     &  0.901	& 0.774	 & 0.430	& 0.185	 & 0.463  & 1.114  & \textbf{0.727}  \\
&  \textcolor{blue!75!cyan}{$-3$}    &  0.899	& 0.769	 & 0.423	& 0.182	 & 0.458  & 1.112  & \textbf{0.727}  \\
&  \textcolor{blue!80!black}{$-4$}   &  0.899	& 0.766	 & 0.422	& 0.184	 & 0.458  & 1.113  & \textbf{0.727}  \\
&  \textcolor{blue!65!black}{$-5$}   &  0.897	& 0.763	 & 0.420	& 0.185	 & 0.456  & 1.111  & 0.726  \\
&  \textcolor{blue!40!black}{$-10$}  &  0.885	& 0.747  & 0.422	& 0.188	 & 0.452  & 1.107  & 0.724  \\

\bottomrule % <-- Bottomrule here
\end{tabular}%}
\vspace{-5mm}
\end{center}
\end{table}

% 0.861	& 0.748	 & 0.454	& 0.200  & 0.467  & 1.322  & 0.722  \\
% 0.876	& 0.763	 & 0.462	& 0.198	 & 0.475  & 1.334  & 0.724  \\
% 0.882	& 0.767	 & 0.462	& 0.198	 & 0.476  & 1.340  & 0.724  \\
% 0.888	& 0.771	 & 0.461	& 0.196	 & 0.476  & 1.343  & 0.725  \\
% 0.894	& 0.774	 & 0.457	& 0.192	 & 0.474  & 1.342  & 0.725  \\
% 0.899	& 0.774	 & 0.451	& 0.189	 & 0.471  & 1.335  & 0.726  \\
% 0.901	& 0.775	 & 0.442	& 0.187	 & 0.468  & 1.323  & 0.727  \\
% 0.903	& 0.775	 & 0.436	& 0.186	 & 0.466  & 1.316  & 0.727  \\
% 0.901	& 0.772	 & 0.429	& 0.185	 & 0.462  & 1.315  & 0.727  \\
% 0.900	& 0.768	 & 0.421	& 0.182	 & 0.457  & 1.316  & 0.726  \\
% 0.898	& 0.761	 & 0.417	& 0.183	 & 0.453  & 1.312  & 0.724  \\
% 0.895	& 0.757	 & 0.415	& 0.184	 & 0.452  & 1.310  & 0.724  \\
% 0.887	& 0.745	 & 0.416	& 0.190	 & 0.450  & 1.306  & 0.723  \\

\boldparagraph{Contribution of progress and preference estimation} 
In \tabref{tab:abl_losses}, we start with the BCE-only baseline in the first row.
Our BCE-only baseline already achieves performance comparable to the SOTA model TOP~\cite{zhao2025accident} (\tabref{tab:cap_dada_sota}). In particular, it obtains a similar mAUC (0.419 vs. 0.429) while achieving substantially earlier anticipation in terms of mTTA (1.105 vs. 0.864). 
Adding progress estimation (second row) yields substantial improvements across all metrics, highlighting the benefit of modeling a continuously increasing risk signal. Incorporating preference estimation provides slight additional gains, particularly at the earlier horizon of 1.5 seconds before the accident. The best overall performance is achieved when both are used together, as shown in the last row.

\boldparagraph{Risk functions}
We experiment with different risk functions by varying the $\alpha$ parameter from \eqnref{eq:risk_target_exp}, as illustrated in \figref{fig:risk-targets-wrap}, including a linear baseline. 
Quantitative results on the CAP dataset are reported in \tabref{tab:risk_func_abl}. The linear risk function provides a strong and competitive baseline across metrics. We observe a clear trade-off induced by the shape of the risk function: exponential functions with positive $\alpha$ improve performance at longer anticipation horizons, reflected by higher AUC$_{1.0s}$ and AUC$_{1.5s}$, whereas negative $\alpha$ values yield stronger performance at shorter horizons, particularly on AUC$_{0.0s}$. This provides control over the anticipation behavior of the model. We choose $\alpha=3$, which achieves the best overall performance in terms of both mAUC and mTTA.

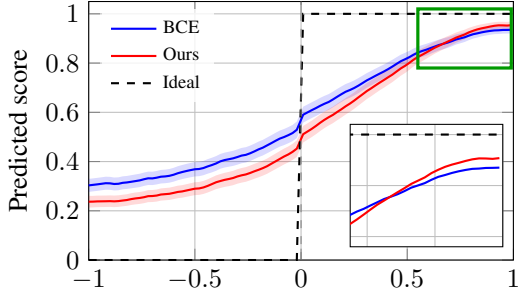
\begin{wrapfigure}{r}{0.5\textwidth}
\centering
\vspace{-10pt}

\begin{tikzpicture}

\begin{axis}[
    name=mainplot,
    width=7.2cm,
    height=5cm,
    xmin=-1, xmax=1,
    ymin=0, ymax=1.05,
    ylabel={Predicted score},
    grid=major,
    xtick={-1,-0.5,0,0.5,1},
    ytick={0,0.2,0.4,0.6,0.8,1.0},
    legend style={
        at={(0.03,0.97)},
        anchor=north west,
        font=\scriptsize,
        draw=none,
        fill=white,
        fill opacity=0.8,
        text opacity=1
    },
]

% Upper and lower invisible boundaries
\addplot[name path=bce_upper, draw=none, forget plot]
table[x=x, y expr=\thisrow{y}+\thisrow{std}/10, col sep=comma]
{figures/our_metric/tikz_data_cap/only_bce_loss.csv};

\addplot[name path=bce_lower, draw=none, forget plot]
table[x=x, y expr=\thisrow{y}-\thisrow{std}/10, col sep=comma]
{figures/our_metric/tikz_data_cap/only_bce_loss.csv};

% Std shaded region
\addplot[blue, opacity=0.15, forget plot]
fill between[of=bce_upper and bce_lower];

% Main curve
\addplot[blue, thick]
table[x=x, y=y, col sep=comma]
{figures/our_metric/tikz_data_cap/only_bce_loss.csv};
\addlegendentry{BCE}

% \addplot[orange, thick]
% table[x=x, y=y, col sep=comma] {figures/our_metric/tikz_data_cap/bce_+_margin_loss.csv};

% \addplot[green!60!black, thick]
% table[x=x, y=y, col sep=comma] {figures/our_metric/tikz_data_cap/bce_+_progress_loss.csv};

\addplot[name path=ours_upper, draw=none, forget plot]
table[x=x, y expr=\thisrow{y}+\thisrow{std}/10, col sep=comma]
{figures/our_metric/tikz_data_cap/bce_+_progress_+_margin_loss.csv};

\addplot[name path=ours_lower, draw=none, forget plot]
table[x=x, y expr=\thisrow{y}-\thisrow{std}/10, col sep=comma]
{figures/our_metric/tikz_data_cap/bce_+_progress_+_margin_loss.csv};

\addplot[red, opacity=0.15, forget plot]
fill between[of=ours_upper and ours_lower];

\addplot[red, thick]
table[x=x, y=y, col sep=comma]
{figures/our_metric/tikz_data_cap/bce_+_progress_+_margin_loss.csv};
\addlegendentry{Ours}

\addplot[black, dashed, thick]
table[x=x, y=ideal, col sep=comma] {figures/our_metric/tikz_data_cap/only_bce_loss.csv};
\addlegendentry{Ideal}

\addplot[gray, dashed] coordinates {(0,0) (0,1.05)};

\node[anchor=north, font=\small, align=center] at (axis cs:0,0) [yshift=-15pt] {Anomaly\\appear};
\node[anchor=north, font=\small, align=center] at (axis cs:-1,0) [yshift=-15pt] {Video\\start};
\node[anchor=north, font=\small, align=center] at (axis cs:1,0) [yshift=-15pt] {Accident\\occur};

% zoom region box on main plot
\draw[green!60!black, very thick] % rounded corners
    (axis cs:0.55,0.78) rectangle (axis cs:0.99,1.02);

\end{axis}

% inset zoom plot
\begin{axis}[
    at={(mainplot.south east)},
    anchor=south east,
    xshift=-4pt,
    yshift=5pt,
    width=3.6cm,
    height=3.2cm,
    xmin=0.55, xmax=1.0,
    ymin=0.78, ymax=1.02,
    grid=major,
    axis background/.style={fill=white},
    tick label style={font=\scriptsize},
    ticks=none,
]

\addplot[blue, thick]
table[x=x, y=y, col sep=comma] {figures/our_metric/tikz_data_cap/only_bce_loss.csv};

% \addplot[orange, thick]
% table[x=x, y=y, col sep=comma] {figures/our_metric/tikz_data_cap/bce_+_margin_loss.csv};

% \addplot[green!60!black, thick]
% table[x=x, y=y, col sep=comma] {figures/our_metric/tikz_data_cap/bce_+_progress_loss.csv};

\addplot[red, thick]
table[x=x, y=y, col sep=comma] {figures/our_metric/tikz_data_cap/bce_+_progress_+_margin_loss.csv};

\addplot[black, dashed, thick]
table[x=x, y=ideal, col sep=comma] {figures/our_metric/tikz_data_cap/only_bce_loss.csv};

\end{axis}

\end{tikzpicture}
% \vspace{-15pt}

\caption{\textbf{Separation profile on CAP.} BCE inflates pre-onset scores, whereas our model provides a smoother rise and improved accuracy near the accident, as shown zoomed-in.}
\label{fig:sep_profil_cap}
    
% \vspace{-5mm}

\end{wrapfigure}

\boldparagraph{Separation score} 
In \figref{fig:sep_profil_cap}, we plot the normalized mean risk scores of the BCE-only baseline and our model on the CAP subset. The curves are centered at 0, corresponding to anomaly onset, and normalized to $[-1,1]$, representing the start of the video and the moment of the accident, respectively. Ideally, an accident anticipation model should assign low scores before anomaly onset and then gradually increase them toward the accident. 
The BCE-only model predicts higher scores before anomaly onset, indicating increased false alarm spikes during safe-driving segments. In contrast, our model exhibits a more controlled increase after onset and achieves higher scores closer to the accident, as highlighted in the zoomed-in region.

\begin{wraptable}{l}{0.48\textwidth}
\vspace{-3mm}
\caption{\textbf{Separation scores.}}
\vspace{-2mm}
\begin{center}
\resizebox{0.45\columnwidth}{!}{
\begin{tabular}{llccc}
\toprule % <-- Toprule here
& Model & $s_{\text{pre}}$ $\downarrow$ & $s_{\text{post}}$ $\uparrow$  & $s$ $\uparrow$ \\
\midrule % <-- Midrule here

\multirow{2}{*}{\rotatebox{90}{\scriptsize CAP}} & BCE    & 0.420 & \textbf{0.808} & 0.694   \\
& \ours       & \textbf{0.344} & 0.792 & \textbf{0.724}  \\

\midrule % <-- Midrule here

\multirow{2}{*}{\rotatebox{90}{\scriptsize DADA}} & BCE    & 0.372 & 0.763 & 0.696   \\
& \ours       & \textbf{0.319} & \textbf{0.798} & \textbf{0.740} \\

\bottomrule % <-- Bottomrule here
\end{tabular}}
\end{center}
\label{tab:separation_score_ablations}
\vspace{-7mm}
\end{wraptable}

In \tabref{tab:separation_score_ablations}, we report separation scores on CAP and DADA. Our model consistently assigns lower $s_{\text{pre}}$ scores on both datasets, indicating fewer false positives in the safe-driving zone. Although the BCE-only obtains a slightly higher $s_{\text{post}}$ on CAP, its substantially higher $s_{\text{pre}}$ leads to a lower final score. In contrast, our model achieves cleaner separation between $s_{\text{pre}}$ and $s_{\text{post}}$, resulting in consistently higher Separation Score $s$ on both datasets.

\section{Conclusion}
\label{sec:conc}

We reformulated accident anticipation from binary classification to continuous risk estimation by modeling risk as a temporally evolving signal that increases as the crash approaches. To better capture global model behavior, we additionally introduced the Separation Score, which evaluates predicted risk curves beyond local temporal windows. Our method significantly advances SOTA on the CAP and DADA subsets of MM-AU, and outperforms the Nexar challenge winner. We identified that most gains come from explicit continuous risk progression estimation, while preference-based ranking provides additional benefits for longer-horizon anticipation. 
Our ablations on risk functions further reveal a trade-off between short- and long-horizon anticipation behavior. 
Finally, our Separation Score reveals important differences in false alarm behavior and pre-accident risk evolution that are not captured by existing metrics.

\boldparagraph{Limitations} A key limitation of our work is that we study accident anticipation as a video monitoring problem, whereas real-world driving safety also depends on factors such as driver state, vehicle dynamics, and sensor reliability. In addition, our CAP-to-Nexar transfer experiments highlight the challenges of cross-domain generalization, which should be studied more systematically given the difficulty of collecting large-scale real crash data.

\section*{Acknowledgment}
Eray Çakar and Fatma Güney are funded by the European Union (ERC, ENSURE, 101116486). %Nermin Samet is partially funded by the Horizon Europe project ELLIOT (GA No. 101214398).

\bibliography{refs, related_work}
\bibliographystyle{unsrtnat}

%%%%%%%%%%%%%%%%%%%%%%%%%%%%%%%%%%%%%%%%%%%%%%%%%%%%%%%%%%%%

\newpage

\appendix

\section{Appendix}
In this appendix, we first provide additional details on our model size and inference speed (\appref{app:model_details}). We then present our additional ablation experiments with varying FPR (\appref{app:varying_fpr}) and backbone (\appref{app:varying_backbone}), followed by Nexar results using standard anticipation metrics (\appref{app:nexar_results}).

\subsection{More details on model}
\label{app:model_details}

Our model, \ours, has 307M total parameters, of which 53M are trainable. 
We train our models on a single NVIDIA A100 GPU for 2000 iter with a batch size of 32.
During inference, our model achieves an FPS of 14.44 on the same GPU. 

Our backbone video foundation model (VideoMAE~\cite{tong2022videomae}) takes, by default, $16$ frames as input. In order to feed $5$ frames (\ie 0.5 second clips), we devised a padding scheme. We repeat each frame 3 times, except for the last one, which we repeat 4 times.

\subsection{Complementary results with varying FPR}
\label{app:varying_fpr}

So far, we have evaluated our method at FPR $\leq 0.1$. In \tabref{tab:abl_fpr}, we further report performance under both stricter (FPR $\leq 0.01$) and more lenient (FPR $\leq 1$) constraints. When the FPR is constrained to low values, performance drops substantially, particularly for long-horizon predictions. Nevertheless, the mTTA results indicate that our method can still issue warnings within a reasonable time window, achieving above $0.8$\,s on both datasets even under the strictest setting.

%%%% Table CAP Results.
\begin{table}[h]
\caption{\textbf{Performance across varying FPR.} Despite drops at low FPR, our method maintains timely warnings, achieving $0.8$\,s mTTA at FPR $\leq 0.01$.}
\label{tab:abl_fpr}

\begin{center}
% \resizebox{1.0\columnwidth}{!}{
\begin{tabular}{llcccccc}
\toprule % <-- Toprule here
{Dataset} & $\lambda$ & AUC$_{0.0s}^{\lambda}$ &  AUC$_{0.5s}^{\lambda}$ & AUC$_{1.0s}^{\lambda}$ &  AUC$_{1.5s}^{\lambda}$ & mAUC$^{\lambda}$ & 
 mTTA$^{\lambda}$ \\
\midrule % <-- Midrule here

\multirow{3}{*}{CAP~\cite{fang2022cognitive}}
  & 0.01 &   0.721 & 0.500 & 0.106 & 0.025 & 0.210 & 0.817 \\
  & 0.1  &   0.887 & 0.775	& 0.465	& 0.201	& 0.481 & 1.125 \\
  & 1.0  &   0.976 & 0.945 & 0.852 & 0.701 & 0.833 & 1.820 \\
\midrule

\multirow{3}{*}{DADA~\cite{fang2021dada}}
  & 0.01  &    0.732	& 0.503 & 0.136	& 0.038	& 0.226  & 0.826 \\
  & 0.1   &    0.861	& 0.751	& 0.398	& 0.170	& 0.440  & 1.069 \\
  & 1.0   &    0.960	& 0.922	& 0.782	& 0.631	& 0.778  & 1.597 \\

\bottomrule % <-- Bottomrule here
\end{tabular}%}

%\vspace{-5mm}
\end{center}
\end{table}

%%%% Table CAP Results.
\begin{table}
\caption{\textbf{Varying backbone.} Video models (V-JEPAv2 and VideoMAE) clearly outperform the image backbone (DINOv2), emphasizing spatio-temporal features; similar results across video encoders show backbone-agnostic behavior for our method.}
\label{tab:abl_backbones}

\begin{center}
\resizebox{1.0\columnwidth}{!}{
\begin{tabular}{llcccccc}
\toprule % <-- Toprule here
{Dataset} & Backbone & AUC$_{0.0s}^{0.1}$ &  AUC$_{0.5s}^{0.1}$ & AUC$_{1.0s}^{0.1}$ &  AUC$_{1.5s}^{0.1}$ & mAUC$^{0.1}$ & 
 mTTA$^{0.1}$  \\
\midrule % <-- Midrule here

\multirow{3}{*}{CAP~\cite{fang2022cognitive}}
  & DINOv2~\cite{oquab2023dinov2} &  0.837	& 0.573	& 0.272	& 0.136  & 0.327 & 0.721 \\
  & V-JEPAv2~\cite{assran2025v}  &   0.881	& 0.778	& 0.426	& 0.176  & 0.460 & 1.120 \\
  & VideoMAE~\cite{tong2022videomae} & \textbf{0.887} & \textbf{0.775} & \textbf{0.465} & \textbf{0.201} & \textbf{0.481} &\textbf{1.125} \\

\midrule

\multirow{3}{*}{DADA~\cite{fang2021dada}}
  & DINOv2~\cite{oquab2023dinov2}  &  0.674 & 0.358 & 0.179 & 0.117  & 0.218  & 0.850 \\
  & V-JEPAv2~\cite{assran2025v}   &   0.596	& 0.516	& 0.314	& \textbf{0.177} & 0.336 & {1.011} \\
  & VideoMAE~\cite{tong2022videomae}   &   \textbf{0.861} & \textbf{0.751} & \textbf{0.398} & 0.170 & \textbf{0.440} & \textbf{1.069} \\

\bottomrule % <-- Bottomrule here
\end{tabular}}
\end{center}
\end{table}

\subsection{Experiments with varying backbones}
\label{app:varying_backbone}

In \tabref{tab:abl_backbones}, we evaluate our method using an image foundation model and an alternative video foundation model. For the image model, we use DINOv2~\cite{oquab2023dinov2} with GRU~\cite{chung2014empirical} layers to incorporate temporal information. For the video model, we replace the VideoMAE \cite{wang2023videomae} backbone with V-JEPA~\cite{assran2025v}. All variants are trained with the same hyperparameters, as described in \secref{sec:setup}. As expected, the image backbone performs substantially worse than both video backbones, highlighting the importance of explicit spatio-temporal representations. V-JEPA achieves comparable performance to our model and even outperforms it in mTTA on DADA, indicating that our method is not tied to a specific video encoder.

\subsection{Complementary results on Nexar challenge}
\label{app:nexar_results}

%%%% Table Nexar Results.
\begin{table}[h]
\caption{Results using standard metrics on the Nexar~\cite{moura2025nexar} dataset.}
\label{tab:nexar_results}

\begin{center}
%\resizebox{1.0\columnwidth}{!}{
\begin{tabular}{lcccccccc}
\toprule % <-- Toprule here
{Method}  & AP$_{0.5s}$ & AP$_{1.0s}$ &  AP$_{1.5s}$  & mAP & mAUC &mAUC$^{0.1}$ &  mTTA$^{0.1}$ & $s$ \\
\midrule % <-- Midrule here
    
% Winner   & - &  - & - & 0.885  & - & - & - \\
% Runner-up   & - &  - & -  & 0.875& - & - & - \\
% % 3rd   & - &  - & -  & 0.854& - & - & - \\
% % 4rd   & - &  - & -  & 0.823& - & - & - \\
% TOP & - &  - & -  & 0.909 & - & - & - \\

% \midrule % <-- Midrule here

{\ours} & 0.919 & 0.911 & 0.854 & 0.895 & 0.895 & 0.510 & 1.285 & 0.665 \\

\bottomrule % <-- Bottomrule here
\end{tabular}%}

% \begin{tablenotes} %[para]
%     \item 
%     \footnotesize Results are taken from \url{https://www.kaggle.com/competitions/nexar-collision-prediction/leaderboard}.
% \end{tablenotes}

\end{center}
\end{table}

In the main text, we present the Nexar challenge results using the only official metric, mAP. In \cref{tab:nexar_results}, we share the results using standard metrics, AP for different horizons, mAUC, mTTA, and Separation Score.

Nexar~\cite{moura2025nexar} dataset differs from MM-AU~\cite{fang2024abductive} in two main ways. First, the positive videos (\ie videos with accident) in the test set are trimmed either 0.5, 1.0 or 1.5 seconds before the accident happens. Hence, they do not include the accident frames. 
Second, the dataset misses the ``anomaly onset" labels for the positive videos in the test set. 

To calculate the mTTA metric (recall that it requires a $t_{ai}$ annotation), we artificially set the ``anomaly onset" label to the same point as $t_h$. Specifically, we set $t_{ai}$ exactly 2.0 seconds before the accident happens.

%%%%%%%%%%%%%%%%%%%%%%%%%%%%%%%%%%%%%%%%%%%%%%%%%%%%%%%%%%%%

\end{document}